\documentclass{article}

\usepackage[preprint]{neurips_2026}

\usepackage[utf8]{inputenc} 
\usepackage[T1]{fontenc}    
\usepackage{hyperref}       
\usepackage{url}            
\usepackage{booktabs}       
\usepackage{amsfonts}       
\usepackage{nicefrac}       
\usepackage{microtype}      
\usepackage{xcolor}         
\usepackage{graphicx}
\usepackage{amsmath} 
\usepackage{mathrsfs} 
\usepackage{bm} 
\usepackage{amsthm}
\usepackage{enumitem}
\usepackage{algorithm}
\usepackage{algpseudocode}
\usepackage{multirow} 
\usepackage[table]{xcolor}
\usepackage{amsmath, amssymb}
\usepackage{listings}

\newtheorem{proposition}{Proposition}
\title{IRIS: Interpolative Rényi Iterative Self-play for Large Language Model Fine-Tuning}

\author{%
  Wenjie Liao\textsuperscript{1}\thanks{Corresponding author.} \quad
  Like Wu\textsuperscript{1} \quad
  Liangjie Zhao\textsuperscript{2} \quad
  Shihui Xu\textsuperscript{1} \quad
  Shigeru Fujimura\textsuperscript{1} \\[6pt]
  \textsuperscript{1}Graduate School of Information, Production and Systems, Waseda University, Fukuoka, Japan \\
  \textsuperscript{2}Institute of Computing Technology, University of the Chinese Academy of Sciences \\[4pt]
  \texttt{\{jie3040@aoni, wulike@fuji\}.waseda.jp, zhaoliangjie55@gmail.com} \\
  \texttt{shxu@toki.waseda.jp, fujimura@waseda.jp} \\
}

\begin{document}

\maketitle

\begin{abstract}
Self-play fine-tuning enables large language models to improve beyond supervised fine-tuning without additional human annotations by contrasting annotated responses with self-generated ones. Many existing methods rely on a fixed divergence regime. SPIN is closely related to a KL-based regime, SPACE to a Jensen-Shannon-style objective via noise contrastive estimation, and SPIF to $\chi^2$-regularized self-play. Since these divergences exhibit different strengths depending on the distributional gap between model and target, no single choice appears to provide favorable learning dynamics across training stages. We propose IRIS (Interpolative R\'enyi Iterative Self-play), a R\'enyi-based self-play fine-tuning framework with a continuously adjustable objective. IRIS decomposes into two independent tilted risk terms over annotated and synthetic data, with exponential importance weights controlled by the order parameter~$\alpha$. We show that several self-play objectives can be interpreted as limiting or representative regimes at particular values of~$\alpha$, providing a unified theoretical perspective on these methods. An adaptive order schedule further adjusts~$\alpha$ to the distributional gap, shifting from sharper importance weighting early in training to smoother refinement near convergence. Theoretically, we establish the fixed-point property of IRIS and analyze how~$\alpha$ controls gradient concentration. Experiments on Zephyr-7B and Qwen2.5-3B across ten benchmarks show that IRIS improves upon baselines, reaching 44.57\% average score with gains across iterations. In our setting, IRIS with only 26$k$ annotated samples surpasses standard supervised fine-tuning trained on the full 200$k$ dataset.
\end{abstract}

\section{Introduction}

Large language models (LLMs) have achieved remarkable capabilities through post-training alignment with human preferences\cite{mishra2022cross,thoppilan2022lamda,chung2024scaling,roziere2023code,ouyang2022training,openai,anthropic_claude35sonnet_2024a,google_gemini25_2025}. The prevailing approaches, including reinforcement learning from human feedback (RLHF)\cite{christiano2017deep,azar2024general,xiong2023iterative,bai2022training} and direct preference optimization (DPO)\cite{rafailov2023direct,tu2025enhancing,wu2024self}, rely on curated preference datasets that are expensive to acquire and difficult to scale\cite{dong2024abilities,liu2023makes,lee2024rlaif,yuan2024self}. This practical constraint motivates a natural question: can an LLM continue to improve beyond the ceiling of supervised fine-tuning (SFT)\cite{ouyang2022training,tunstall2023zephyr} without any additional human-annotated data? Self-play fine-tuning (SPIN)\cite{chen2024self} provides an affirmative answer. Drawing inspiration from the self-play mechanism in AlphaGo Zero\cite{silver2017mastering}, SPIN frames fine-tuning as a two-player game in which the current model learns to distinguish human-written responses from those generated by its previous iteration. Empirically, SPIN enables a weak SFT model to match models trained with DPO without acquiring new human annotations, and has been applied to various LLM fields\cite{yuan2024self2,wu2024self,gao2024sprec}.

Despite this conceptual elegance, SPIN exhibits several practical limitations that have motivated a growing body of follow-up work\cite{liao2026drift,wang2025space,wang2026triplets,wu2026self,li2026your,yin2024sapo,ji2024selfplay,ding2024sail,ye2024online}. The gap-based objective, which contrasts real and synthetic responses through their reward difference, degenerates to a trivial constant as the two distributions converge during training. This gradient vanishing phenomenon causes instability and performance degradation in later iterations. SPACE\cite{wang2025space} addresses this problem by reformulating the objective via noise contrastive estimation (NCE)\cite{gutmann2010noise,gutmann2012noise}, replacing the gap-based loss with independent classification losses for real and synthetic samples. From a complementary perspective, SGALM\cite{wu2026self} constructs a GAN-style adversarial game\cite{goodfellow2014generative} within a single LLM, arriving at a structurally similar binary classification framework. Meanwhile, T-SPIN\cite{wang2026triplets} tackles the vanishing reward advantage across iterations by introducing triplets that incorporate historical model outputs alongside current ones, maintaining a persistent learning signal through the entire training process. DRIFT\cite{liao2026drift} observes that SPIN applies uniform training pressure across all tokens, including those the model already predicts correctly, and proposes difference-aware masking to concentrate the gradient on error regions. Most recently, SPIF\cite{li2026your} reveals that SPIN's implicit reward function can grow unboundedly, and introduces $\chi^2$ divergence\cite{pearson1900criterion,neyman1949contribution} regularization to constrain reward magnitudes.

Each of these methods remedies a specific failure mode, but we observe that they share a common root cause. Many recent self-play fine-tuning methods\cite{chen2024self, wang2025space, wang2026triplets, liao2026drift, wu2026self, li2026your, alami2024regularization, wu2024sppo, tang2025rspo, yuan2024selfrewarding} can be interpreted through the lens of $f$-divergence optimization, although they differ in data construction, optimization parameterization, and auxiliary design. The existing strategies fall into three categories accordingly. The first retains the gap-based contrastive structure under the KL divergence, strengthening it via historical anchoring, token-level masking, or opponent regularization\cite{chen2024self,wang2026triplets,liao2026drift,alami2024regularization}. The second replaces the gap with independent binary classification of real and synthetic samples under the Jensen-Shannon divergence, encompassing noise contrastive estimation, adversarial, and self-rewarding approaches\cite{wang2025space,wu2026self,yuan2024selfrewarding}. The third enforces explicit divergence constraints or game-theoretic equilibrium formulations grounded in $\chi^2$ regularization or Nash convergence\cite{li2026your,wu2024sppo,tang2025rspo}. This taxonomy suggests a practical limitation of existing methods: a single fixed divergence may not provide equally favorable learning dynamics across all stages of training. The KL divergence provides smooth gradients when the model is far from the target but loses sensitivity as the distributions converge. The $\chi^2$ divergence offers bounded rewards and fine-grained discrimination near convergence, yet over-penalizes outliers early on. The JS divergence ensures stability through independent optimization but is less sensitive to differences in later stages. This explains why each category excels in certain regimes while poorly in others. The comparisons are shown in Figure 1.

\begin{figure*}[!t]
    \centering
    \includegraphics[width=\textwidth]{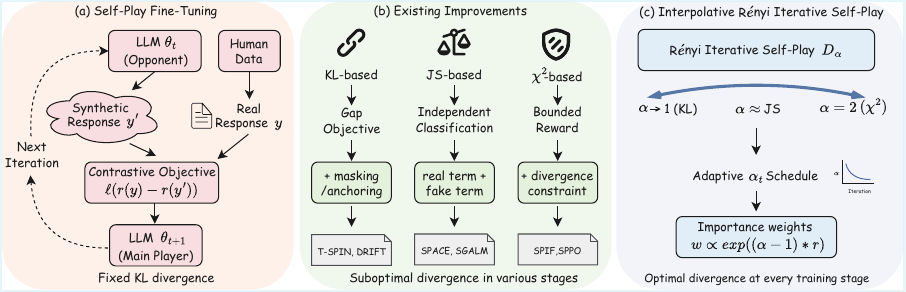}
    \caption{Overview of self-play fine-tuning paradigms: (a)~the standard framework trains $\pi_{\theta_{t+1}}$ to distinguish human from synthetic responses via a contrastive objective under a fixed KL divergence; (b)~existing improvements adopt KL-based, JS-based, or $\chi^2$-based objectives, each suboptimal outside its preferred training regime; (c)~IRIS places these methods within a R\'enyi-based perspective, with an adaptive schedule $\alpha_t$ that interpolates among corresponding divergence regimes during training.}
    \label{fig:1}
\end{figure*}

Building on this insight, we propose \textbf{IRIS} (\textbf{I}nterpolative \textbf{R}\'enyi \textbf{I}terative \textbf{S}elf-play), a unified framework that resolves the divergence selection problem through the R\'enyi divergence\cite{renyi1961measures}. The R\'enyi divergence $D_\alpha(p\|q)$ is a one-parameter generalization that continuously interpolates among the KL divergence ($\alpha \to 1$), the $\chi^2$ divergence ($\alpha = 2$), and the Hellinger distance ($\alpha = 1/2$)~\cite{vanerven2014renyi,cichocki2010families}, covering several divergence regimes closely related to those underlying existing self-play objectives. Exploiting the variational representation of this family~\cite{li2016renyi,birrell2022variational}, we derive the IRIS objective as a loss that decomposes into two independent terms over real and synthetic data, inheriting the stability of independent classification. The gradient features exponential importance weights $\tilde{w} \propto \exp((\alpha\!-\!1) \cdot r)$ that continuously concentrate learning on informative samples through a principled, tunable mechanism. We further introduce an adaptive order schedule $\alpha_t$ that adjusts the divergence geometry to the distributional gap: larger $\alpha$ encourages mode-covering exploration early in training, while smaller $\alpha$ enables mode-seeking refinement near convergence, selecting the optimal divergence at every stage.

Theoretically, we establish the fixed-point property of IRIS, analyze its gradient structure, and show that several self-play objectives can be interpreted as representative or limiting regimes under particular choices of $\alpha$. We further introduce an adaptive schedule to adjust the divergence geometry during training. Empirically, experiments on Zephyr-7B\cite{tunstall2023zephyr} and Qwen2.5-3B-Instruct\cite{yang2025qwen3} across ten tasks from the Open LLM Leaderboard\cite{beeching_openllmleaderboard_2023} show that IRIS reaches 44.57\% average score at Iter~4, with up to 10.75-point gains over SFT on tasks such as IFEval\cite{zhou2023instruction}. IRIS maintains stable improvements across all four iterations, whereas several strong baselines plateau or show less stable behavior later.

\textbf{Our contributions.} (i) We present a R\'enyi-based perspective that connects representative self-play objectives through distinct divergence regimes; (ii) We propose IRIS, a self-play fine-tuning framework with a continuously adjustable objective and an adaptive order schedule; (iii) We analyze the fixed-point and gradient properties of IRIS; (iv) Experiments across ten benchmarks show that IRIS outperforms other self-play baselines, while maintaining stable gains across iterations.

\section{Preliminaries}

In this section, we introduce the necessary background on supervised fine-tuning and self-play fine-tuning, and examine existing improvements through the unifying lens of $f$-divergence theory.

\noindent \textbf{Supervised Fine-Tuning.} Supervised fine-tuning (SFT) is a straightforward yet powerful approach for large language models to achieve effective task adaptation using labeled data\cite{dong2024abilities,wang2023far,tan2024large}. Given a prompt $\mathbf{x}$ sampled from a task-specific distribution $q(\cdot)$ and a corresponding high-quality response $\mathbf{y}$ from the target distribution $p_{\mathrm{data}}(\cdot|\mathbf{x})$, supervised fine-tuning (SFT) learns a policy $p_\theta$ by maximizing the log-likelihood of annotated data:
\begin{equation}\label{eq:sft}
\mathcal{L}_{\mathrm{SFT}}(\theta) = \mathbb{E}_{\mathbf{x} \sim q(\cdot),\, \mathbf{y} \sim p_{\mathrm{data}}(\cdot|\mathbf{x})} \big[ \log p_\theta(\mathbf{y}|\mathbf{x}) \big].
\end{equation}
Maximizing \eqref{eq:sft} is equivalent to minimizing the KL divergence $\mathrm{KL}\!\left(p_{\mathrm{data}}(\cdot|\mathbf{x}) \,\|\, p_\theta(\cdot|\mathbf{x})\right)$. By the non-negativity of the KL divergence, the optimal solution satisfies $p_{\theta^*}\!(\cdot|\mathbf{x}) = p_{\mathrm{data}}(\cdot|\mathbf{x})$. However, approaching this optimum requires extensive annotated data, limiting practical applicability of SFT.

\noindent \textbf{Self-Play Fine-Tuning.} Self-play fine-tuning\cite{chen2024self,gao2024sprec,ren2024learning} addresses the data scarcity limitation of SFT by augmenting training with model-generated responses. The framework consists of a main player and an opponent player, both instantiated from the same LLM with different parameters. At each iteration $t+1$, the opponent player $p_{\theta_t}$ generates a synthetic response $\mathbf{y}' \sim p_{\theta_t}(\cdot|\mathbf{x})$ for each prompt $\mathbf{x}$. The main player then updates its parameters to distinguish the annotated response $\mathbf{y}$ from the synthetic one $\mathbf{y}'$. Defining the log-ratio reward and objectives as
\begin{equation}\label{eq:reward}
r(\mathbf{x}, \hat{\mathbf{y}}) = \log p_\theta(\hat{\mathbf{y}}|\mathbf{x}) - \log p_{\theta_t}(\hat{\mathbf{y}}|\mathbf{x}),
\end{equation}
\begin{equation}\label{eq:spin}
\mathcal{L}_{\mathrm{SPIN}}(\theta) = \mathbb{E}\!\left[\ell\!\left(r(\mathbf{x}, \mathbf{y}) - r(\mathbf{x}, \mathbf{y}')\right)\right].
\end{equation}
where $\ell(\cdot)$ is a monotonically decreasing convex function, typically chosen as the logistic loss $\ell(t) = \log(1 + e^{-t})$, and the expectation is taken over $\mathbf{x} \sim q(\cdot)$, $\mathbf{y} \sim p_{\mathrm{data}}(\cdot|\mathbf{x})$, and $\mathbf{y}' \sim p_{\theta_t}(\cdot|\mathbf{x})$. The global optimum of \eqref{eq:spin} is attained when $p_\theta = p_{\mathrm{data}}$, at which point the reward gap vanishes for all prompt-response pairs and the objective degenerates into a constant independent of $\theta$. This fixed-point property guarantees convergence in principle, but it also implies that the training signal weakens as $p_{\theta_t}$ approaches $p_{\mathrm{data}}$, causing optimization instability in later iterations.

\noindent \textbf{Self-Play Improvements as $f$-Divergence Minimization.} The instability described above has motivated a growing body of improvements to the self-play framework~\cite{wang2025space,wang2026triplets,liao2026drift,wu2026self,li2026your,alami2024regularization,wu2024sppo,tang2025rspo}.A closer examination reveals that these methods, despite their diverse
motivations, share a common mathematical structure rooted in
$f$-divergence theory~\cite{ali1966general,csiszar1967information}. Recall
that an $f$-divergence $D_f(p\|q) = \mathbb{E}_q[f(p/q)]$ is uniquely
determined by its convex generator $f$, and admits a variational
representation via the Fenchel conjugate~\cite{nguyen2010estimating} that
enables tractable optimization through neural network
parameterization~\cite{nowozin2016fgan}. Through this lens, existing self-play methods cluster into three families
according to the divergence they implicitly minimize. The first family
operates under the KL divergence, preserving the gap-based contrastive
structure of SPIN while strengthening it through historical
anchoring~\cite{wang2026triplets}, token-level
masking~\cite{liao2026drift}, or opponent
regularization~\cite{alami2024regularization}. The second family
corresponds to the Jensen-Shannon
divergence~\cite{goodfellow2014generative}, replacing the reward gap with
independent binary classification of real and synthetic samples via noise
contrastive estimation~\cite{wang2025space} or adversarial
training~\cite{wu2026self}. The third family employs the $\chi^2$
divergence, enforcing explicit divergence
constraints~\cite{li2026your} or game-theoretic equilibrium
formulations~\cite{wu2024sppo,tang2025rspo} that bound the reward
magnitude.

This classification exposes a fundamental limitation shared by all
existing approaches: each operates under a single fixed divergence
throughout training. As discussed in Section~1, no single $f$-divergence
provides uniformly optimal learning dynamics across all stages, since
different divergences exhibit complementary strengths depending on the
distributional gap between the model and the target. This observation
motivates a framework capable of adapting its divergence behavior during
training, which we develop in the next section using the R\'enyi
divergence family\cite{renyi1961measures,vanerven2014renyi,li2016renyi,
birrell2022variational,cichocki2010families,mironov2017renyi,
amari2016information,gil2013renyi,wang2024beyond}.
\section{Method}

In this section, we present IRIS, a unified self-play fine-tuning framework built on the R\'enyi divergence. We construct the IRIS objective, analyze its gradient structure, and introduce an adaptive order schedule throughout training. Additional proofs and derivatives are provided in Appendix B.

\subsection{The IRIS Objective} 

The analysis in Section II reveals that each existing self-play method commits to a single $f$-divergence throughout training, and no fixed divergence provides uniformly optimal learning dynamics across all stages. To address this limitation, we seek a parameterized objective that continuously interpolates among different divergence behaviors while preserving the desirable properties of self-play fine-tuning: a fixed point at $p_\theta = p_{\mathrm{data}}$, independent optimization over real and synthetic data, and tractable gradient computation. Our method builds on the \emph{tilted risk} principle~\cite{li2021tilted}. For a random variable $Z$ under distribution $p$, the tilted risk of order $s \neq 0$ is defined as
\begin{equation}\label{eq:tilted}
\mathcal{R}_s(Z;\, p) = \frac{1}{s} \log \mathbb{E}_{p}\!\left[e^{s Z}\right].
\end{equation}
This quantity is the scaled cumulant generating function of $Z$, interpolating between the expectation ($s \to 0$) and the essential supremum ($s \to +\infty$). The tilted risk is connected to the R\'enyi divergence: for any distributions $p$ and $q$, the R\'enyi divergence of order $\alpha$ satisfies $D_\alpha(p\|q) = \mathcal{R}_\alpha\!\left(\log\frac{p}{q};\, q\right)$~\cite{vanerven2014renyi}, establishing the tilted risk as a building block for R\'enyi-based objectives. Using the log-ratio reward $r(\mathbf{x}, \hat{\mathbf{y}}) = \log p_\theta(\hat{\mathbf{y}}|\mathbf{x}) - \log p_{\theta_t}(\hat{\mathbf{y}}|\mathbf{x})$ defined in~\eqref{eq:reward}, we construct the IRIS objective for the main player by applying tilted risks of complementary orders to the real and synthetic data:
\begin{equation}\label{eq:iris}
\mathcal{L}_{\mathrm{IRIS}}(\theta;\, \alpha) = \underbrace{-\frac{1}{\alpha\!-\!1}\, \log \mathbb{E}_{\mathbf{y} \sim p_{\mathrm{data}}}\!\left[e^{(\alpha-1)\, r(\mathbf{x}, \mathbf{y})}\right]}_{\text{tilted reward on real data}} \;+\; \underbrace{\frac{1}{\alpha}\, \log \mathbb{E}_{\mathbf{y}' \sim p_{\theta_t}}\!\left[e^{\alpha\, r(\mathbf{x}, \mathbf{y}')}\right]}_{\text{tilted penalty on synthetic data}},
\end{equation}
where the outer expectation over prompts $\mathbf{x} \sim q(\cdot)$ is omitted for notational clarity, and $\alpha > 0$ with $\alpha \neq 1$ is the R\'enyi order parameter. The first term is the negated tilted risk of order $(\alpha\!-\!1)$ evaluated on real data, which encourages the model to assign high reward to human-annotated responses with sensitivity controlled by $\alpha$. The second term is the tilted risk of order $\alpha$ on synthetic data, which penalizes high reward assigned to model-generated responses. The two terms operate independently on real and synthetic samples, inheriting the structural stability of independent classification~\cite{wang2025space,wu2026self}. The asymmetry in orders $(\alpha\!-\!1)$ and $\alpha$ arises naturally from the R\'enyi variational principle~\cite{birrell2022variational} and ensures that the two terms remain balanced across the full range of $\alpha$. A requirement for any self-play objective is that the target distribution $p_{\mathrm{data}}$ constitutes a fixed point. We establish this for IRIS.

\begin{proposition}[Fixed Point]\label{prop:fixed}
For any $\alpha > 0$ with $\alpha \neq 1$, the global minimum of $\mathcal{L}_{\mathrm{IRIS}}(\theta;\, \alpha)$ over $\theta$ is attained at $p_\theta = p_{\mathrm{data}}$. At this optimum,
\begin{equation}\label{eq:opt_value}
\mathcal{L}_{\mathrm{IRIS}}(\theta^*;\, \alpha) = -\frac{1}{\alpha}\, D_\alpha(p_{\mathrm{data}} \| p_{\theta_t}),
\end{equation}
where $D_\alpha$ denotes the R\'enyi divergence of order $\alpha$. Furthermore, if $p_{\theta_t} = p_{\mathrm{data}}$, then $\mathcal{L}_{\mathrm{IRIS}}(\theta^*;\, \alpha) = 0$, confirming that the equilibrium $p_\theta = p_{\theta_t} = p_{\mathrm{data}}$ is a fixed point of the iterative self-play process.
\end{proposition}

\noindent \textbf{Gradient Analysis and Importance Weighting.} To understand the optimization dynamics of IRIS, we derive the gradient of~\eqref{eq:iris} with respect to $\theta$ and analyze how order $\alpha$ controls the model training.

\begin{proposition}[Gradient Structure]\label{prop:gradient}
The gradient of $\mathcal{L}_{\mathrm{IRIS}}(\theta;\, \alpha)$ with respect to $\theta$ takes the form
\begin{equation}\label{eq:grad}
\nabla_\theta \mathcal{L}_{\mathrm{IRIS}} = -\mathbb{E}_{\mathbf{y} \sim p_{\mathrm{data}}}\!\left[\tilde{w}^{+}\!(\mathbf{y})\, \nabla_\theta \log p_\theta(\mathbf{y}|\mathbf{x})\right] + \mathbb{E}_{\mathbf{y}' \sim p_{\theta_t}}\!\left[\tilde{w}^{-}\!(\mathbf{y}')\, \nabla_\theta \log p_\theta(\mathbf{y}'|\mathbf{x})\right],
\end{equation}
where the normalized importance weights are
\begin{equation}\label{eq:weights}
\tilde{w}^{+}\!(\mathbf{y}) = \frac{e^{(\alpha-1)\, r(\mathbf{x}, \mathbf{y})}}{\mathbb{E}_{p_{\mathrm{data}}}\!\left[e^{(\alpha-1)\, r}\right]}, \qquad \tilde{w}^{-}\!(\mathbf{y}') = \frac{e^{\alpha\, r(\mathbf{x}, \mathbf{y}')}}{\mathbb{E}_{p_{\theta_t}}\!\left[e^{\alpha\, r}\right]}.
\end{equation}
\end{proposition}

The gradient in~\eqref{eq:grad} admits a natural interpretation as importance-weighted likelihood updates: real responses are reinforced while synthetic ones are suppressed, with exponential weights acting as a differentiable attention mechanism over samples. For $\alpha > 1$, $\tilde{w}^{+} \propto \exp((\alpha\!-\!1) r)$ emphasizes real responses that already receive high reward, whereas $\tilde{w}^{-} \propto \exp(\alpha r)$ concentrates on synthetic responses the model incorrectly favors, steering updates toward the most informative samples on both sides. The order $\alpha$ thereby governs the sharpness of reweighting, smoothly interpolating between uniform gradients at $\alpha \to 1$ and highly concentrated ones at large $\alpha$, and yields a continuously adjustable soft counterpart to hard selection schemes. This exponential tilting further connects IRIS to tilted empirical risk minimization~\cite{li2021tilted}, where $\alpha$ balances average-case and worst-case performance.

\noindent \textbf{Unification of Existing Methods.} A notable consequence of the tilted risk construction is that the IRIS objective recovers the divergence families underlying existing self-play methods at specific values of $\alpha$, confirming that these methods were implicitly operating within the R\'enyi family. The independent treatment of real and synthetic terms in~\eqref{eq:iris} further parallels the decoupled structure of SPACE~\cite{wang2025space} and SGALM~\cite{wu2026self} under the Jensen-Shannon divergence. The correspondences of various regimes are summarized in Table~\ref{tab:unify} and three informative regimes are listed below:

\begin{itemize}[leftmargin=1.2em,itemsep=0pt,topsep=2pt,parsep=0pt]
\item \textbf{KL regime} ($\alpha \to 1$). Using the limit $\frac{1}{s}\log\mathbb{E}[e^{sZ}] \to \mathbb{E}[Z]$ as $s \to 0$, $\mathcal{L}_{\mathrm{IRIS}}$ converges to $-\mathbb{E}_{p_{\mathrm{data}}}[r] + \log\mathbb{E}_{p_{\theta_t}}[e^{r}]$, the Donsker-Varadhan form of the KL divergence~\cite{donsker1983asymptotic}, and the importance weights become uniform. This recovers the divergence regime of SPIN~\cite{chen2024self} together with its gap-based variants~\cite{wang2026triplets,liao2026drift,alami2024regularization}, which approximate this KL objective via logistic losses.
\item \textbf{$\chi^2$ regime} ($\alpha = 2$). The identity $\mathbb{E}_{p_{\theta_t}}[e^{2r}] = 1 + \chi^2(p_\theta\|p_{\theta_t})$ implies that the second term directly penalizes the $\chi^2$ divergence between the model and the opponent, which aligns with the bounded-reward formulations of SPIF~\cite{li2026your}, SPPO~\cite{wu2024sppo}, and RSPO~\cite{tang2025rspo}.
\item \textbf{Hellinger regime} ($\alpha = 1/2$). Under the squared Hellinger distance, $\tilde{w}^{+}\!\propto e^{-r/2}$ emphasizes underperforming real responses, producing mode-covering behavior absent from prior methods.
\end{itemize}

\begin{table}[t]
\centering
\caption{Unification of self-play fine-tuning methods through the IRIS framework. Representative self-play objectives can be related to specific regimes within the R\'enyi divergence family.}
\label{tab:unify}
\vspace{2pt}
\small
\begin{tabular}{llll}
\toprule
Order $\alpha$ & Divergence & Weight profile & Corresponding methods \\
\midrule
$\alpha \to 1$ & KL & Uniform & SPIN\cite{chen2024self}, T-SPIN\cite{wang2026triplets}, DRIFT\cite{liao2026drift} \\
$\alpha = 1$  (NCE) & Jensen-Shannon & Sigmoid & SPACE\cite{wang2025space}, SGALM\cite{wu2026self} \\
$\alpha = 2$ & $\chi^2$ & Quadratic & SPIF\cite{li2026your}, SPPO\cite{wu2024sppo} \\
$\alpha = 1/2$ & Hellinger & Inverse-exponential & (new regime via IRIS) \\
\bottomrule
\end{tabular}
\end{table}

\subsection{Adaptive Order Schedule}\label{sec:schedule}

The unification established above motivates a natural question: rather than committing to a single $\alpha$ throughout training, can we adapt $\alpha_t$ at each iteration? This question is central to IRIS, because the appropriate divergence geometry depends on the current distributional gap between the policy and the target. We therefore consider two complementary strategies for choosing the order during training.

\noindent \textbf{Distributional Gap Feedback.} We set $\alpha_t$ based on the estimated divergence between the model and target distributions:
\begin{equation}\label{eq:adapt}
\alpha_t = 1 + c \cdot \hat{D}_t, \qquad \hat{D}_t = \mathbb{E}_{\mathbf{y} \sim p_{\mathrm{data}}}\!\left[r_t(\mathbf{x}, \mathbf{y})\right] - \mathbb{E}_{\mathbf{y}' \sim p_{\theta_t}}\!\left[r_t(\mathbf{x}, \mathbf{y}')\right],
\end{equation}
where $c > 0$ is a scaling constant and $\hat{D}_t$ is the expected reward gap at the beginning of iteration $t$, serving as a proxy for the distributional distance\cite{chen2024self}. This choice ties the schedule directly to the training signal already used by self-play, and therefore avoids introducing an additional external criterion for tuning $\alpha_t$. When the model is far from the target ($\hat{D}_t$ large), $\alpha_t$ is large, promoting mode-covering exploration through the $\chi^2$-like behavior and concentrated importance weights. This places more emphasis on informative high-gap samples, which is beneficial when the model and data distributions remain substantially separated. As the model converges ($\hat{D}_t \to 0$), $\alpha_t$ decreases toward $1$, transitioning to the KL regime that provides smoother, more uniform gradients near equilibrium.

\noindent \textbf{Geometric Annealing.} For simplicity and reproducibility, we also consider a predetermined schedule:
\begin{equation}\label{eq:anneal}
\alpha_t = \alpha_{\max} \cdot \left(\frac{\alpha_{\min}}{\alpha_{\max}}\right)^{t/T},
\end{equation}
which interpolates from $\alpha_{\max}$ to $\alpha_{\min}$ over $T$ iterations. This schedule requires no statistics and is straightforward to implement. Both strategies share the same inductive bias: larger $\alpha$ early for exploration through importance weights, and smaller $\alpha$ later for refinement through uniform gradients. In Section IV, we compare both strategies and find that distributional gap feedback outperforms geometric annealing, as it responds to dynamics rather than following a predetermined trajectory.

\subsection{Algorithm Summary}\label{sec:algo}

We summarize IRIS in Algorithm~\ref{alg:iris}. The procedure follows the standard self-play pipeline~\cite{chen2024self}, with two modifications: (i) the loss function in~\eqref{eq:iris} replaces the gap-based or fixed-divergence objectives of prior methods, and (ii) the order $\alpha_t$ is updated at each iteration based on reward statistics. In practice, the log-sum-exp operations in~\eqref{eq:iris} are stabilized by subtracting the batch maximum before exponentiation~\cite{blanchard2021accurately}, the importance weights $\tilde{w}^{\pm}$ in~\eqref{eq:weights} are normalized within each mini-batch to prevent any single sample from dominating the gradient, and the reward gap $\hat{D}_t$ in~\eqref{eq:adapt} is estimated from the first mini-batch of each iteration at negligible additional cost. IRIS is fully compatible with existing self-play codebases~\cite{chen2024self,wang2025space}, requiring only a substitution of the loss computation module while leaving the generation, data processing, and distributed training components unchanged. 

\begin{algorithm}[t]
\caption{IRIS: Interpolative R\'enyi Iterative Self-Play}
\label{alg:iris}
\begin{algorithmic}[1]

\State \textbf{Input:} SFT model $p_{\theta_0}$, dataset $\mathcal{D} = \{(\mathbf{x}_i, \mathbf{y}_i)\}_{i=1}^n$, iterations $T$, constant $c$

\For{$t = 0, 1, \ldots, T\!-\!1$}
    \State Generate synthetic responses $\mathbf{y}'_i \sim p_{\theta_t}(\cdot|\mathbf{x}_i)$
    \State Compute reward statistics and set $\alpha_t$ via~\eqref{eq:adapt} or~\eqref{eq:anneal}
    \State Update $\theta_{t+1} \leftarrow \arg\min_\theta \mathcal{L}_{\mathrm{IRIS}}(\theta;\, \alpha_t)$
\EndFor

\State \textbf{Output:} Fine-tuned model $p_{\theta_T}$

\end{algorithmic}
\end{algorithm}

\section{Experiments}\label{sec:exp}

In this section, we conduct extensive experiments to evaluate the effectiveness of IRIS. We first describe the experimental settings, and then report the main results with analyses. Finally, we present ablation studies on the adaptive order schedule. More details are provided in Appendix C.

\subsection{Experimental Setup}\label{sec:setup}

\noindent \textbf{Models and training.} Following SPIN~\cite{chen2024self} and subsequent works~\cite{wang2025space,wang2026triplets,li2026your}, we randomly sample 50$k$ prompt-response pairs from the first round of interactions in Ultrachat200k~\cite{ding2023enhancing} and evaluate on Zephyr-7B-SFT-Full~\cite{tunstall2023zephyr} and Qwen2.5-3B-Instruct~\cite{yang2025qwen3}. At each of five self-play iterations (Iter~0 through Iter~4), we generate one synthetic response per prompt and combine it with the annotated response for training. We use RMSProp~\cite{hinton2012neural} as optimizer with a global batch size of 64 and 2 epochs per iteration. For the adaptive schedule in~\eqref{eq:adapt}, we set $c = 0.5$ with $\alpha_t \in [0.5, 3.0]$. All the experiments run on 8 NVIDIA H100 GPUs using the Alignment Handbook~\cite{tunstall2024alignment} and Accelerate~\cite{gugger2022accelerate}.

\begin{table}[t]
\centering
\caption{Performance (\%) comparisons on Zephyr-7B, where we report improvements ({\color[rgb]{0,0.5,0}green}) or degradations ({\color{red}red}) over the previous iteration. \textbf{Bold}: best result across all iterations for each method.}
\label{tab:main}
\vspace{2pt}
\resizebox{\textwidth}{!}{%
\begin{tabular}{rc ccc ccc ccc c c}
\toprule
\multicolumn{2}{c}{\multirow{2}{*}{\textbf{Model}}} & \multicolumn{3}{c}{Math \& Logic} & \multicolumn{3}{c}{Multi-Domain Knowledge} & \multicolumn{3}{c}{Commonsense Reasoning} & IF & \multirow{2}{*}{\textbf{Avg}} \\
\cmidrule(lr){3-5} \cmidrule(lr){6-8} \cmidrule(lr){9-11} \cmidrule(lr){12-12}
& & GSM8K & MATH & MUSR & MMLU & MMLUP & GPQA & HellaSwag & WG & BBH & IFEval & \\
\midrule
\multicolumn{2}{c}{Zephyr-7B} & 25.73 & 1.71 & 38.07 & 56.44 & 29.19 & 28.80 & 81.89 & 74.20 & 44.54 & 2.97 & 38.35 \\
\multicolumn{2}{c}{SFT} & 41.06 & 3.15 & 39.79 & 57.22 & 29.23 & 28.27 & 83.40 & 73.08 & 44.61 & 20.07 & 41.99 \\
\midrule
\multirow{5}{*}{\rotatebox[origin=c]{90}{SPIN}}
& Iter0 & 30.25 & 4.30 & 41.42 & 56.43 & 28.24 & 29.31 & 84.17 & 73.63 & 44.53 & 8.47 & 40.08 \\
& Iter1 & 32.68 & 2.87 & 38.79 & 56.51 & 27.72 & 29.35 & 83.63 & 73.84 & 44.72 & 7.63 & 39.77\,{\scriptsize\color{red}($-$0.31)} \\
& Iter2 & 36.85 & 2.72 & 41.53 & 57.46 & 28.53 & 28.61 & 83.58 & 73.90 & 43.78 & 14.72 & \textbf{41.17}\,{\scriptsize\color[rgb]{0,0.5,0}($+$1.40)} \\
& Iter3 & 33.54 & 2.58 & 39.13 & 55.71 & 25.83 & 27.54 & 83.41 & 74.35 & 43.22 & 22.34 & 40.77\,{\scriptsize\color{red}($-$0.40)} \\
& Iter4 & 31.63 & 2.43 & 38.58 & 54.82 & 25.17 & 26.93 & 83.20 & 74.21 & 42.96 & 21.87 & 40.19\,{\scriptsize\color{red}($-$0.58)} \\
\midrule
\multirow{5}{*}{\rotatebox[origin=c]{90}{SPACE}}
& Iter0 & 40.72 & 3.67 & 40.15 & 57.60 & 28.94 & 28.83 & 83.38 & 73.71 & 43.14 & 10.82 & 41.10 \\
& Iter1 & 40.83 & 4.20 & 40.67 & 58.17 & 29.08 & 29.15 & 83.52 & 73.56 & 43.31 & 12.29 & 41.48\,{\scriptsize\color[rgb]{0,0.5,0}($+$0.38)} \\
& Iter2 & 41.15 & 4.48 & 41.23 & 58.24 & 29.15 & 29.27 & 83.47 & 73.82 & 43.62 & 14.36 & 41.88\,{\scriptsize\color[rgb]{0,0.5,0}($+$0.40)} \\
& Iter3 & 41.37 & 4.64 & 41.59 & 58.12 & 29.23 & 29.41 & 83.55 & 73.93 & 43.87 & 16.43 & 42.21\,{\scriptsize\color[rgb]{0,0.5,0}($+$0.33)} \\
& Iter4 & 41.82 & 4.73 & 41.86 & 58.07 & 29.30 & 29.58 & 83.61 & 73.67 & 44.18 & 17.12 & \textbf{42.40}\,{\scriptsize\color[rgb]{0,0.5,0}($+$0.19)} \\
\midrule
\multirow{5}{*}{\rotatebox[origin=c]{90}{T-SPIN}}
& Iter0 & 36.47 & 3.40 & 37.28 & 56.53 & 29.07 & 28.23 & 82.82 & 73.61 & 44.28 & 7.31 & 39.90 \\
& Iter1 & 39.82 & 3.53 & 38.01 & 57.02 & 29.28 & 29.14 & 83.08 & 73.85 & 44.63 & 26.73 & 42.51\,{\scriptsize\color[rgb]{0,0.5,0}($+$2.61)} \\
& Iter2 & 40.93 & 3.77 & 38.68 & 57.12 & 29.40 & 29.59 & 83.24 & 73.78 & 44.62 & 27.83 & 42.90\,{\scriptsize\color[rgb]{0,0.5,0}($+$0.39)} \\
& Iter3 & 40.85 & 3.96 & 39.32 & 57.83 & 29.52 & 30.27 & 83.42 & 73.86 & 45.14 & 29.39 & 43.36\,{\scriptsize\color[rgb]{0,0.5,0}($+$0.46)} \\
& Iter4 & 40.40 & 3.98 & 39.84 & 57.76 & 29.47 & 30.92 & 83.28 & 73.93 & 45.08 & 30.84 & \textbf{43.56}\,{\scriptsize\color[rgb]{0,0.5,0}($+$0.20)} \\
\midrule
\multirow{5}{*}{\rotatebox[origin=c]{90}{SPIF}}
& Iter0 & 33.81 & 4.14 & 41.73 & 56.82 & 28.63 & 29.45 & 84.25 & 73.47 & 44.68 & 9.43 & 40.64 \\
& Iter1 & 37.60 & 4.38 & 41.83 & 57.35 & 29.06 & 29.57 & 83.86 & 73.58 & 44.73 & 14.92 & 41.69\,{\scriptsize\color[rgb]{0,0.5,0}($+$1.05)} \\
& Iter2 & 39.42 & 4.53 & 42.19 & 57.68 & 29.36 & 29.84 & 83.67 & 73.74 & 44.76 & 18.83 & \textbf{42.40}\,{\scriptsize\color[rgb]{0,0.5,0}($+$0.71)} \\
& Iter3 & 38.85 & 4.47 & 41.58 & 57.26 & 29.10 & 29.53 & 83.54 & 73.62 & 44.38 & 19.42 & 42.18\,{\scriptsize\color{red}($-$0.22)} \\
& Iter4 & 38.27 & 4.30 & 41.14 & 56.93 & 28.87 & 29.29 & 83.42 & 73.53 & 44.15 & 19.17 & 41.91\,{\scriptsize\color{red}($-$0.27)} \\
\midrule
\rowcolor{red!6}
& Iter0 & 35.18 & 3.63 & 41.87 & 56.78 & 29.12 & 29.52 & 84.13 & 73.68 & 44.81 & 10.29 & 40.90 \\
\rowcolor{red!6}
& Iter1 & 41.53 & 4.17 & 42.08 & 57.86 & 29.67 & 30.14 & 83.95 & 73.92 & 45.23 & 22.61 & 43.12\,{\scriptsize\color[rgb]{0,0.5,0}($+$2.22)} \\
\rowcolor{red!6}
& Iter2 & 41.74 & 4.50 & 42.21 & 58.34 & 30.12 & 30.58 & 83.87 & 74.16 & 45.67 & 26.49 & 43.77\,{\scriptsize\color[rgb]{0,0.5,0}($+$0.65)} \\
\rowcolor{red!6}
& Iter3 & 41.90 & 4.83 & 42.78 & 58.72 & 30.43 & 30.70 & 83.93 & 74.28 & 45.82 & 29.36 & 44.28\,{\scriptsize\color[rgb]{0,0.5,0}($+$0.51)} \\
\rowcolor{red!6}
\multirow{-5}{*}{\rotatebox[origin=c]{90}{\textbf{IRIS}}}
& Iter4 & 42.21 & 4.97 & 42.89 & 58.93 & 30.57 & 30.96 & 84.05 & 74.43 & 45.82 & 30.82 & \textbf{44.57}\,{\scriptsize\color[rgb]{0,0.5,0}($+$0.29)} \\
\bottomrule
\end{tabular}%
}
\end{table}

\begin{figure}[!h]
    \centering
    \includegraphics[width=\textwidth]{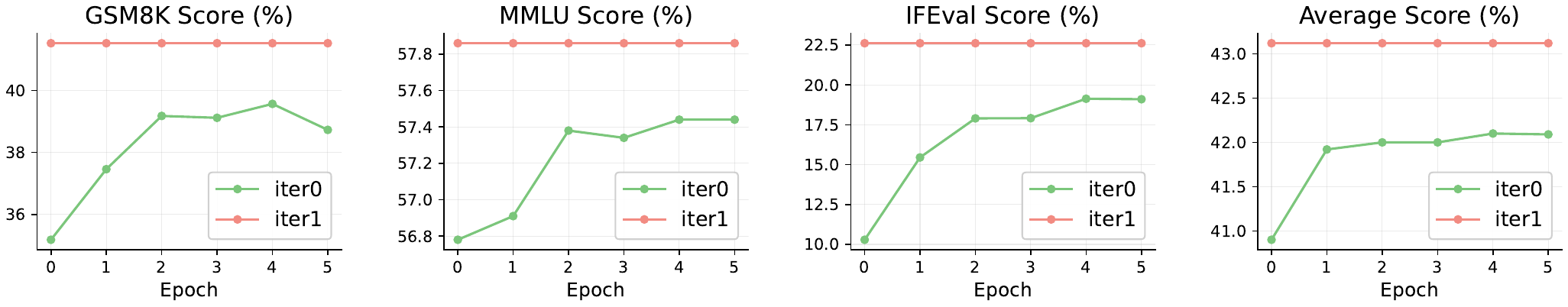}
    \caption{Performance comparison between training at iteration~0 for multiple epochs on fixed synthetic responses ({\color[rgb]{0,0.5,0}green}) and training at iteration~1 for two epochs on regenerated responses ({\color{red}red}).}
    \label{fig:epochs}
\end{figure}

\noindent \textbf{Evaluation and baselines.} We evaluate on ten diverse tasks from the HuggingFace Open LLM Leaderboard~\cite{beeching_openllmleaderboard_2023}, grouped into mathematical and logical reasoning (GSM8K~\cite{cobbe2021training}, MATH~\cite{hendrycks2021math}, MUSR~\cite{sprague2024musr}), multi-domain knowledge (MMLU~\cite{hendrycks2021measuring}, MMLU-Pro~\cite{wang2024mmlupro}, GPQA~\cite{rein2023gpqa}), commonsense reasoning (HellaSwag~\cite{zellers2019hellaswag}, WinoGrande~\cite{sakaguchi2021winogrande}, BBH~\cite{suzgun2022challenging}), and instruction following (IFEval~\cite{zhou2023instruction}), using the Language Model Evaluation Harness~\cite{gao2023framework}. We compare IRIS against SFT trained on the full 200$k$ dataset, SPIN~\cite{chen2024self}, SPACE~\cite{wang2025space}, T-SPIN~\cite{wang2026triplets}, and SPIF~\cite{li2026your}. All self-play methods are trained on the same 50$k$ subset with identical generation and training configurations for fair comparison.

\subsection{Main Results}\label{sec:main_results}

\noindent \textbf{Comparison across methods.} Table~\ref{tab:main} presents the main results on Zephyr-7B across all ten benchmarks. Several observations emerge from these results. First, all self-play methods surpass both the pretrained model and SFT, confirming the effectiveness of iterative self-play even with limited annotated data. Second, IRIS consistently achieves the highest average score across all iterations, reaching 44.57\% at Iter~4, compared to 43.56\% for T-SPIN, 42.40\% for SPACE and SPIF, and 41.17\% for SPIN. Third, the improvements from IRIS are particularly pronounced on MATH, MUSR and IFEval tasks with 1.82-point, 3.10-point and 10.75-point increases respectively compared with SFT base, where the adaptive divergence schedule provides the greatest benefit. Results on Qwen2.5-3B-Instruct, which also exhibit similar overall trends, are provided in Appendix C.3.

\noindent \textbf{Comparison to training with more epochs.}
We examine whether the iterative gains of self-play can be matched by simply training longer
on fixed synthetic responses, extending iteration~0 to five epochs and comparing against
iteration~1 with two epochs on regenerated responses (Figure~\ref{fig:epochs}). The average
score plateaus near 42.1 after five epochs, remaining well below the 43.1 that iteration~1
achieves with two epochs alone, with especially large gaps on GSM8K and IFEval. The
bottleneck lies in the stagnation of negative samples: as the policy improves, the fixed
synthetic responses grow increasingly uninformative. Regenerating responses at iteration~1
supplies negatives that reflect the model's current weaknesses, and in IRIS additionally
triggers a recalibration of~$\alpha_t$ via the adaptive schedule~\eqref{eq:adapt}, keeping the
objective aligned with the evolving distributional gap. These results confirm that data
regeneration, not prolonged training on static data, drives cross-iteration improvement.

\begin{figure}[t]
    \centering
    \includegraphics[width=\textwidth]{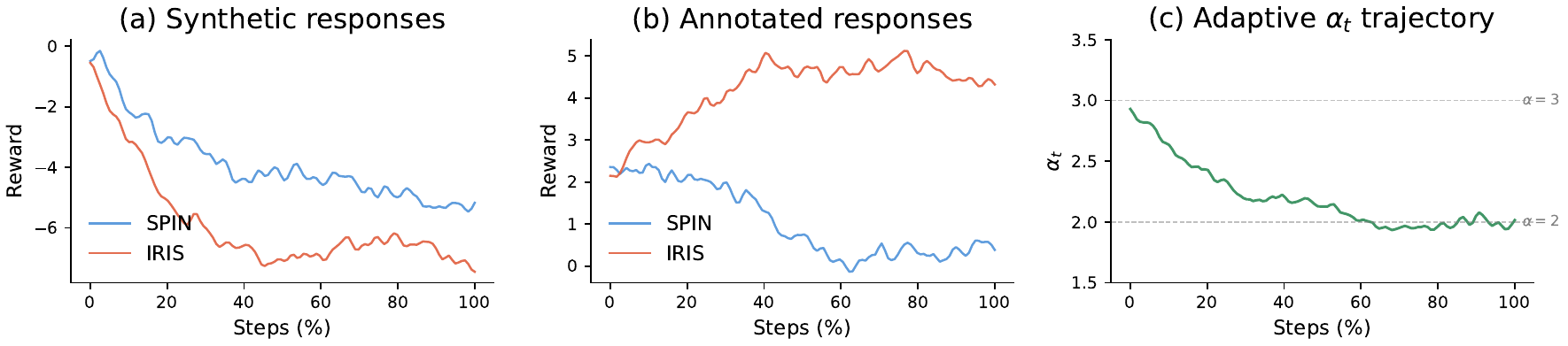}
    \caption{Training dynamics at Iter~0 on Zephyr-7B: (a)~reward of synthetic responses; (b)~reward of annotated responses; (c)trajectory of the adaptive order $\alpha_t$ under distributional gap feedback schedule.}
    \label{fig:dynamics}
\end{figure}

\noindent \textbf{Training dynamics.} Reward decline on high-quality responses is a known issue in preference alignment~\cite{feng2024towards,pal2024smaug,rafailov2024from,xiao2024cal} and was first studied in self-play fine-tuning by SPACE~\cite{wang2025space}. Figure~\ref{fig:dynamics} compares IRIS and SPIN at Iter~0. Panels (a) and (b) show that SPIN enlarges the reward margin, but the reward on annotated responses decreases during training, while the reward on synthetic responses decreases faster. This is a limitation of gap-based optimization: SPIN improves the objective through the relative difference $r(\mathbf{x},\mathbf{y}) - r(\mathbf{x},\mathbf{y}')$, without directly increasing the absolute reward on real data. As a result, a larger margin does not necessarily mean better modeling of high-quality responses. In contrast, IRIS increases the reward on real data and decreases it on synthetic data, since its objective optimizes both terms separately via a tilted reward on annotated data and a tilted penalty on synthetic ones. The margin in IRIS reflects progress toward the target distribution. Panel (c) shows under distributional gap feedback, $\alpha_t$ starts near 3.0 and approaches 2.0, indicating a shift from sharper reweighting early in training to smoother refinement, with training cost recorded in Appendix C.2.

\begin{table}[t]
\centering
\caption{Ablation on order schedule evaluated on Zephyr-7B. All variants use IRIS objective in~\eqref{eq:iris} and differ in how $\alpha_t$ is determined. The Distributional Gap Feedback schedule achieves best performance.}
\label{tab:ablation}
\vspace{2pt}
\small
\begin{tabular}{lccccccc}
\toprule
Schedule strategy & $\alpha$ range & Iter0 & Iter1 & Iter2 & Iter3 & Iter4 & Best Avg \\
\midrule
Fixed $\alpha = 0.5$ & $\{0.5\}$ & 40.21 & 41.08 & 41.39 & 41.53 & 41.67 & 41.67 \\
Fixed $\alpha = 1.5$ & $\{1.5\}$ & 41.17 & 42.46 & 43.03 & 43.14 & 42.91 & 43.14 \\
Fixed $\alpha = 2.0$ & $\{2.0\}$ & 41.58 & 42.73 & 42.86 & 42.51 & 42.18 & 42.86 \\
\midrule
Geometric Annealing & $[0.8, 2.5]$ & 41.36 & 42.89 & 43.47 & 43.78 & 43.92 & 43.92 \\
\rowcolor{red!6}
Gap Feedback (Ours) & $[0.5, 3.0]$ & 40.90 & 43.12 & 43.77 & 44.28 & 44.57 & \textbf{44.57} \\
\bottomrule
\end{tabular}
\end{table}

\subsection{Ablation Studies}\label{sec:ablation}

\noindent \textbf{Effect of the adaptive order schedule.} Table~\ref{tab:ablation} compares five strategies for setting the R\'enyi order~$\alpha$ on Zephyr-7B. Fixing~$\alpha$ at 0.5, 1.5, or 2.0 throughout training corresponds to the Hellinger, intermediate, and $\chi^2$ regimes, respectively. The results reveal a clear tradeoff between stability and aggressiveness: $\alpha = 0.5$ produces the most stable trajectory with no degradation across iterations but converges slowly, $\alpha = 2.0$ delivers the strongest initial gains but becomes unstable after Iter~2, and $\alpha = 1.5$ balances the two yet still plateaus at a lower final performance level. Geometric annealing from $\alpha_{\max} = 2.5$ to $\alpha_{\min} = 0.8$ surpasses all fixed choices by combining early exploration with late refinement within each iteration. The distributional gap feedback schedule in~\eqref{eq:adapt} consistently achieves the highest score at every iteration beyond Iter~0, as it calibrates the order to the actual divergence between the policy and the data distribution rather than following a predetermined decay.

\begin{figure}[t]
    \centering
    \includegraphics[width=\textwidth]{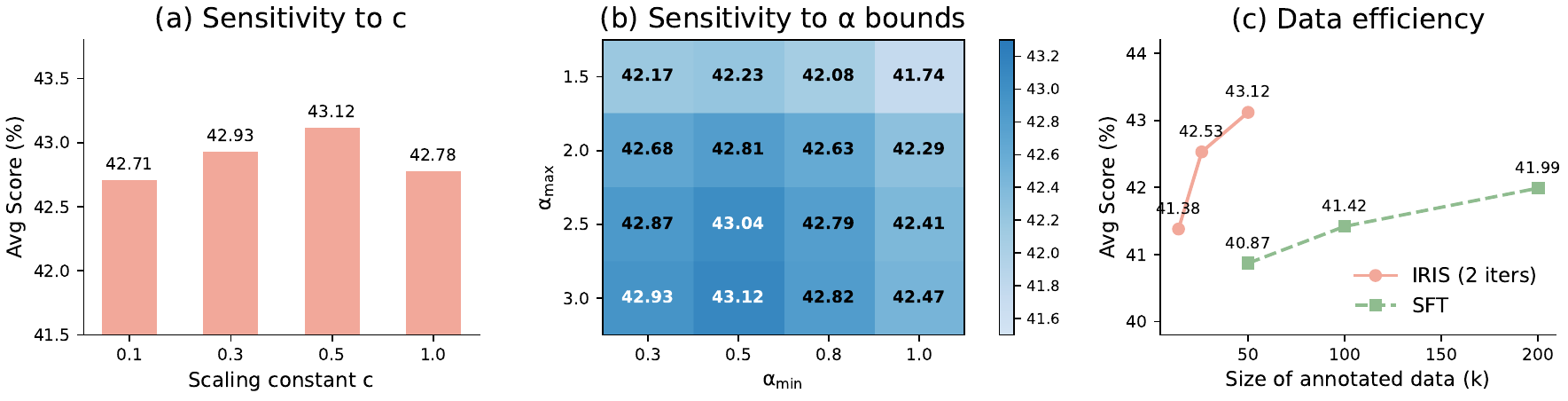}
    \caption{Ablation analysis on Zephyr-7B: (a)~sensitivity to scaling constant $c$ in the distributional gap feedback schedule, evaluated at Iter~1; (b)~heatmap of average scores for different clipping bounds $\alpha_{\min}$ and $\alpha_{\max}$; (c)~comparison between IRIS and SFT with varying amounts of annotated data.}
    \label{fig:ablation}
\end{figure}

\noindent \textbf{Sensitivity to hyperparameters.} We examine the sensitivity of IRIS to the scaling constant $c$ in~\eqref{eq:adapt} and the clipping bounds $[\alpha_{\min}, \alpha_{\max}]$, both evaluated at Iter~1 on Zephyr-7B. Figure~\ref{fig:ablation}(a) reports the average score for $c \in \{0.1, 0.3, 0.5, 1.0\}$: all four values fall between 42.71\% and 43.12\%, a range of less than 0.5 points, indicating that the feedback gain has minimal impact on final performance so long as it remains within a reasonable magnitude. Figure~\ref{fig:ablation}(b) presents a heatmap over $\alpha_{\max} \in \{1.5, 2.0, 2.5, 3.0\}$ and $\alpha_{\min} \in \{0.3, 0.5, 0.8, 1.0\}$. Performance exceeds 42.6\% whenever $\alpha_{\max} \geq 2.0$ and $\alpha_{\min} \leq 0.8$, confirming that the schedule tolerates a broad operating region, while the combination $\alpha_{\max} = 3.0$, $\alpha_{\min} = 0.5$ achieves the peak of 43.12\%. Based on these results, we adopt $c = 0.5$, $\alpha_{\max} = 3.0$, and $\alpha_{\min} = 0.5$ as default values throughout all other experiments.

\noindent \textbf{Data efficiency.} Following the protocol of SPACE~\cite{wang2025space} and T-SPIN~\cite{wang2026triplets}, we compare IRIS with SFT under varying amounts of annotated data. We train IRIS on subsets of 14$k$, 26$k$, and 50$k$ samples for two iterations, and SFT on 50$k$, 100$k$, and 200$k$ samples. Figure~\ref{fig:ablation}(c) shows that IRIS with only 26$k$ annotated responses reaches 42.53\% after two iterations, surpassing SFT trained on the full 200$k$ dataset (41.99\%) by 0.54 points while using roughly one eighth of the annotated data. When the annotated set grows to 50$k$, IRIS further improves to 43.12\%, exceeding 200$k$ SFT by 1.13 points. The gap widens as more annotated data becomes available, suggesting that the adaptive schedule extracts progressively richer training signal from larger annotation pools. These results confirm that iterative self-play amplifies the value of limited annotations, and that the adaptive divergence mechanism in IRIS provides complementary gains by tailoring the R\'enyi order to the current policy state at each iteration rather than relying on a fixed and potentially suboptimal divergence measure. 

\section{Conclusion and discussion}

We have presented IRIS, a self-play fine-tuning framework grounded in the R\'enyi divergence family. The IRIS objective decomposes into two independent tilted risk terms over annotated and synthetic data, yielding decoupled optimization and continuously adjustable importance weighting through the order parameter~$\alpha$. We have shown that several representative self-play objectives, including SPIN, SPACE, and SPIF, can be interpreted within the same R\'enyi-based perspective through particular divergence regimes. The adaptive order schedule shifts from sharper importance weighting early in training to smoother refinement near convergence, responding to the observed distributional gap rather than following a fixed trajectory. Experiments on Zephyr-7B and Qwen2.5-3B across ten benchmarks show that IRIS is competitive with and often improves upon strong baselines, while maintaining stable gains across iterations and ultimately reaching 44.57\% average score at Iter~4.

\noindent \textbf{Limitations and Future Work.} Like all existing self-play methods, IRIS assumes a fixed target distribution throughout training. Extending the framework to non-stationary settings where the target evolves through online interaction is a natural direction, and recent advances in game-theoretic alignment via no-regret learning~\cite{zhang2025iterative} and last-iterate convergence guarantees~\cite{wang2025magnetic} provide promising theoretical foundations for such an extension. Our evaluation is restricted to single-turn generation on models up to 7B parameters; applying IRIS to multi-turn dialogue and scaling to larger architectures through parameter-efficient methods~\cite{hu2022lora} remain unexplored. On the theoretical side, the adaptive schedule in~\eqref{eq:adapt} is empirically effective but lacks formal regret bounds characterizing the optimal trajectory of~$\alpha_t$ on the R\'enyi manifold. Finally, combining the R\'enyi family with broader divergence classes~\cite{wang2024beyond} or extending the mechanism to token-level order adaptation may yield further gains.


\bibliographystyle{plain}
\bibliography{main}









\clearpage
\setcounter{page}{1}

\appendix

\renewcommand{\thesection}{\Alph{section}}
\renewcommand{\thesubsection}{\thesection.\arabic{subsection}}


\section{Additional Related Work}\label{app:related}

\subsection{R\'enyi Divergence and Information Geometry}

The R\'enyi divergence~\cite{renyi1961measures} belongs to the broader family of $\alpha$-divergences studied in information geometry~\cite{amari2016information,cichocki2010families}, with van Erven and Harremoe\"s~\cite{vanerven2014renyi} establishing key properties including monotonicity in the order $\alpha$ and continuity at the KL limit. Its variational representation~\cite{birrell2022variational,li2016renyi} enables tractable optimization in high-dimensional settings and has been applied to variational inference through black-box $\alpha$-divergence minimization~\cite{hernandezlobato2016black}, where the order parameter provides a smooth interpolation between mass-covering and zero-forcing inference. Beyond generative modeling, R\'enyi divergence underpins R\'enyi differential privacy~\cite{mironov2017renyi}, which yields tighter composition bounds than standard $(\varepsilon,\delta)$-DP through moment-based accounting, and connects to distributionally robust optimization via tilted empirical risk minimization~\cite{li2021tilted}, where the tilting parameter controls the tradeoff between average-case and worst-case objectives. Gil~\cite{gil2013renyi} and Wang et al.~\cite{wang2024beyond} offer comprehensive treatments of R\'enyi-based estimation and general divergence optimization, respectively. Our work applies the R\'enyi variational principle to self-play fine-tuning, where the order parameter simultaneously governs the divergence geometry and the importance weighting scheme for adaptive sample selection.

\subsection{Preference Optimization for Language Models}

Post-training alignment of LLMs has been extensively studied through preference-based methods. RLHF~\cite{christiano2017deep,ouyang2022training,bai2022training} first trains a reward model on pairwise human preferences and then optimizes the language model policy via proximal policy optimization. Direct preference optimization (DPO)~\cite{rafailov2023direct} bypasses explicit reward modeling by reparameterizing the reward through the policy itself, and has inspired a family of variants that simplify the pipeline further, including IPO~\cite{azar2024general}, SimPO~\cite{meng2024simpo}, KTO~\cite{ethayarajh2024kto}, and ORPO~\cite{hong2024orpo}. Iterative extensions~\cite{xiong2023iterative,tu2025enhancing,wu2024self,zhan2026mathsmith} regenerate preference pairs at each round, sharing the iterative structure of self-play while still relying on a preference oracle for labeling. Self-play fine-tuning~\cite{chen2024self} operates in a distinct setting: it replaces pairwise preference judgments with the binary distinction between expert-annotated and model-generated responses, removing the need for any preference labels. IRIS inherits this label-free property while introducing adaptive divergence control that is absent from both preference-based and existing self-play approaches.

\section{Theoretical Proofs}\label{app:proofs}

In this section, we provide detailed proofs of Propositions~\ref{prop:fixed} and~\ref{prop:gradient} stated in Section~3, followed by rigorous derivations of the three special-case regimes listed in Table~\ref{tab:unify}. Throughout, we condition on a fixed prompt $\mathbf{x}$ and suppress the dependence on $\mathbf{x}$ for notational brevity; all expectations over prompts can be restored by an outer average over $\mathbf{x}\sim q(\cdot)$.

\subsection{Proof of Proposition~\ref{prop:fixed} (Fixed Point)}

We first show that $p_\theta = p_{\mathrm{data}}$ achieves the claimed optimal value, then prove global optimality via H\"older's inequality.

\noindent \textbf{Step 1: Optimal value.}
Define the density ratio $\rho(\mathbf{y}) \triangleq p_\theta(\mathbf{y}|\mathbf{x}) / p_{\theta_t}(\mathbf{y}|\mathbf{x})$, so $r(\mathbf{x},\mathbf{y}) = \log\rho(\mathbf{y})$. At $p_\theta = p_{\mathrm{data}}$, let $\rho^*(\mathbf{y}) = p_{\mathrm{data}}(\mathbf{y}|\mathbf{x}) / p_{\theta_t}(\mathbf{y}|\mathbf{x})$. Substituting into the two terms of~\eqref{eq:iris}:
\begin{align}
A^* &\triangleq \mathbb{E}_{p_{\mathrm{data}}}\!\left[(\rho^*)^{\alpha-1}\right] = \int p_{\mathrm{data}}(\mathbf{y})\,\frac{p_{\mathrm{data}}(\mathbf{y})^{\alpha-1}}{p_{\theta_t}(\mathbf{y})^{\alpha-1}}\,d\mathbf{y} = \int \frac{p_{\mathrm{data}}(\mathbf{y})^{\alpha}}{p_{\theta_t}(\mathbf{y})^{\alpha-1}}\,d\mathbf{y}, \label{eq:Astar}\\[4pt]
B^* &\triangleq \mathbb{E}_{p_{\theta_t}}\!\left[(\rho^*)^{\alpha}\right] = \int p_{\theta_t}(\mathbf{y})\,\frac{p_{\mathrm{data}}(\mathbf{y})^{\alpha}}{p_{\theta_t}(\mathbf{y})^{\alpha}}\,d\mathbf{y} = \int \frac{p_{\mathrm{data}}(\mathbf{y})^{\alpha}}{p_{\theta_t}(\mathbf{y})^{\alpha-1}}\,d\mathbf{y}. \label{eq:Bstar}
\end{align}
Observing $A^* = B^*$, we denote this common value by $M \triangleq \int p_{\mathrm{data}}(\mathbf{y})^{\alpha}\, p_{\theta_t}(\mathbf{y})^{1-\alpha}\,d\mathbf{y}$, which satisfies $\log M = (\alpha-1)\,D_\alpha(p_{\mathrm{data}}\|p_{\theta_t})$ by the definition of the R\'enyi divergence~\cite{vanerven2014renyi}. The objective value at the optimum is therefore
\begin{equation}\label{eq:opt_derive}
\mathcal{L}_{\mathrm{IRIS}}(\theta^*;\alpha) = \left(-\frac{1}{\alpha-1} + \frac{1}{\alpha}\right)\log M = \frac{-1}{\alpha(\alpha-1)}\cdot(\alpha-1)\,D_\alpha(p_{\mathrm{data}}\|p_{\theta_t}) = -\frac{1}{\alpha}\,D_\alpha(p_{\mathrm{data}}\|p_{\theta_t}),
\end{equation}
which establishes~\eqref{eq:opt_value}. When $p_{\theta_t} = p_{\mathrm{data}}$, $D_\alpha(p_{\mathrm{data}}\|p_{\theta_t}) = 0$ for all $\alpha > 0$~\cite{vanerven2014renyi}, so $\mathcal{L}_{\mathrm{IRIS}}(\theta^*;\alpha) = 0$, confirming the fixed point.

\noindent \textbf{Step 2: Global optimality.}
It remains to show $\mathcal{L}_{\mathrm{IRIS}}(\theta;\alpha) \geq \mathcal{L}_{\mathrm{IRIS}}(\theta^*;\alpha)$ for all $\theta$. For a general $p_\theta$, define
\begin{equation}
A(\theta) \triangleq \mathbb{E}_{p_{\mathrm{data}}}\!\left[\rho^{\alpha-1}\right] = \int \frac{p_{\mathrm{data}}(\mathbf{y})\, p_\theta(\mathbf{y})^{\alpha-1}}{p_{\theta_t}(\mathbf{y})^{\alpha-1}}\,d\mathbf{y}, \qquad B(\theta) \triangleq \mathbb{E}_{p_{\theta_t}}\!\left[\rho^{\alpha}\right] = \int \frac{p_\theta(\mathbf{y})^{\alpha}}{p_{\theta_t}(\mathbf{y})^{\alpha-1}}\,d\mathbf{y}. \label{eq:AB}
\end{equation}
We claim the following inequality holds for all $\alpha > 0$ with $\alpha \neq 1$:
\begin{equation}\label{eq:holder_claim}
A(\theta)^{\alpha} \leq M \cdot B(\theta)^{\alpha-1}.
\end{equation}
We prove~\eqref{eq:holder_claim} by applying H\"older's inequality. Define the non-negative functions
\begin{equation}
u(\mathbf{y}) = \left(\frac{p_{\mathrm{data}}(\mathbf{y})^{\alpha}}{p_{\theta_t}(\mathbf{y})^{\alpha-1}}\right)^{\!1/\alpha}, \qquad v(\mathbf{y}) = \left(\frac{p_\theta(\mathbf{y})^{\alpha}}{p_{\theta_t}(\mathbf{y})^{\alpha-1}}\right)^{\!(\alpha-1)/\alpha}.
\end{equation}
A direct computation confirms the pointwise product:
\begin{equation}
u(\mathbf{y})\cdot v(\mathbf{y}) = \frac{p_{\mathrm{data}}(\mathbf{y})\,p_\theta(\mathbf{y})^{\alpha-1}}{p_{\theta_t}(\mathbf{y})^{\alpha-1}},
\end{equation}
so that $\int u\, v\, d\mathbf{y} = A(\theta)$.

\noindent \emph{Case $\alpha > 1$.}\; Applying H\"older's inequality with conjugate exponents $\alpha$ and $\alpha/(\alpha\!-\!1)$ yields
\begin{equation}\label{eq:holder_apply}
A(\theta) = \int u\, v \leq \left(\int u^{\alpha}\right)^{1/\alpha}\!\left(\int v^{\alpha/(\alpha-1)}\right)^{(\alpha-1)/\alpha}\!.
\end{equation}
The two factors evaluate to $\int u^{\alpha} = M$ and $\int v^{\alpha/(\alpha-1)} = B(\theta)$, giving $A(\theta) \leq M^{1/\alpha}\, B(\theta)^{(\alpha-1)/\alpha}$ and therefore $A(\theta)^{\alpha} \leq M\, B(\theta)^{\alpha-1}$.

\noindent \emph{Case $0 < \alpha < 1$.}\; Since $\alpha < 1$, the exponent pair $(\alpha,\,\alpha/(\alpha\!-\!1))$ has $\alpha \in (0,1)$, so the \emph{reverse} H\"older inequality applies, yielding $A(\theta) \geq M^{1/\alpha}\,B(\theta)^{(\alpha-1)/\alpha}$, or equivalently $A(\theta)^{\alpha} \leq M\,B(\theta)^{\alpha-1}$ (the direction reverses because $\alpha < 1$ flips the power).

In both cases, taking logarithms of~\eqref{eq:holder_claim}:
\begin{equation}\label{eq:log_ineq}
\alpha\log A(\theta) \leq \log M + (\alpha-1)\log B(\theta).
\end{equation}
Rearranging and dividing by $\alpha(\alpha-1)$ (positive for $\alpha > 1$, negative for $0 < \alpha < 1$, which flips the inequality appropriately in each case):
\begin{equation}
-\frac{1}{\alpha-1}\log A(\theta) + \frac{1}{\alpha}\log B(\theta) \geq \frac{-\log M}{\alpha(\alpha-1)} = -\frac{1}{\alpha}\,D_\alpha(p_{\mathrm{data}}\|p_{\theta_t}).
\end{equation}
The left side is exactly $\mathcal{L}_{\mathrm{IRIS}}(\theta;\alpha)$ and the right side is $\mathcal{L}_{\mathrm{IRIS}}(\theta^*;\alpha)$, completing the proof that $p_\theta = p_{\mathrm{data}}$ is a global minimizer. \qed

\subsection{Proof of Proposition~\ref{prop:gradient} (Gradient Structure)}

Since the opponent policy $p_{\theta_t}$ is fixed during each iteration, the reward satisfies $\nabla_\theta r(\mathbf{x},\hat{\mathbf{y}}) = \nabla_\theta \log p_\theta(\hat{\mathbf{y}}|\mathbf{x})$. We differentiate the two terms of~\eqref{eq:iris} with respect to $\theta$.

\noindent \textbf{First term.}\; By the chain rule applied to $-\frac{1}{\alpha-1}\log\mathbb{E}_{p_{\mathrm{data}}}[e^{(\alpha-1)r}]$:
\begin{align}
\nabla_\theta\!\left(-\frac{1}{\alpha\!-\!1}\log\mathbb{E}_{p_{\mathrm{data}}}\!\left[e^{(\alpha-1)r}\right]\right) &= -\frac{1}{\alpha\!-\!1}\cdot\frac{\mathbb{E}_{p_{\mathrm{data}}}\!\left[(\alpha\!-\!1)\,e^{(\alpha-1)r}\,\nabla_\theta\log p_\theta\right]}{\mathbb{E}_{p_{\mathrm{data}}}\!\left[e^{(\alpha-1)r}\right]} \notag\\
&= -\mathbb{E}_{p_{\mathrm{data}}}\!\left[\frac{e^{(\alpha-1)r(\mathbf{x},\mathbf{y})}}{\mathbb{E}_{p_{\mathrm{data}}}\!\left[e^{(\alpha-1)r}\right]}\,\nabla_\theta\log p_\theta(\mathbf{y}|\mathbf{x})\right] \notag\\
&= -\mathbb{E}_{p_{\mathrm{data}}}\!\left[\tilde{w}^+\!(\mathbf{y})\,\nabla_\theta\log p_\theta(\mathbf{y}|\mathbf{x})\right]. \label{eq:grad_term1}
\end{align}

\noindent \textbf{Second term.}\; Analogously for $\frac{1}{\alpha}\log\mathbb{E}_{p_{\theta_t}}[e^{\alpha r}]$:
\begin{align}
\nabla_\theta\!\left(\frac{1}{\alpha}\log\mathbb{E}_{p_{\theta_t}}\!\left[e^{\alpha r}\right]\right) &= \frac{1}{\alpha}\cdot\frac{\mathbb{E}_{p_{\theta_t}}\!\left[\alpha\,e^{\alpha r}\,\nabla_\theta\log p_\theta\right]}{\mathbb{E}_{p_{\theta_t}}\!\left[e^{\alpha r}\right]} \notag\\
&= \mathbb{E}_{p_{\theta_t}}\!\left[\frac{e^{\alpha\, r(\mathbf{x},\mathbf{y}')}}{\mathbb{E}_{p_{\theta_t}}\!\left[e^{\alpha r}\right]}\,\nabla_\theta\log p_\theta(\mathbf{y}'|\mathbf{x})\right] \notag\\
&= \mathbb{E}_{p_{\theta_t}}\!\left[\tilde{w}^-\!(\mathbf{y}')\,\nabla_\theta\log p_\theta(\mathbf{y}'|\mathbf{x})\right]. \label{eq:grad_term2}
\end{align}

Summing~\eqref{eq:grad_term1} and~\eqref{eq:grad_term2} yields~\eqref{eq:grad}. The weights $\tilde{w}^+$ and $\tilde{w}^-$ are non-negative and sum to one within each expectation by construction, so they define valid importance distributions. \qed

\subsection{Derivation of Special Cases}\label{app:special}

We rigorously derive the three limiting and boundary regimes stated in Section~3.1, confirming that each corresponds to a known divergence family.

\subsubsection{KL Regime ($\alpha \to 1$)}

The tilted risk $\mathcal{R}_s(Z;p) = \frac{1}{s}\log\mathbb{E}_p[e^{sZ}]$ is the scaled cumulant generating function of $Z$. A standard Taylor expansion of the moment generating function around $s = 0$ gives
\begin{equation}\label{eq:taylor_mgf}
\mathbb{E}_p\!\left[e^{sZ}\right] = 1 + s\,\mathbb{E}_p[Z] + \frac{s^2}{2}\mathbb{E}_p[Z^2] + O(s^3),
\end{equation}
so $\frac{1}{s}\log(1+s\,\mathbb{E}_p[Z]+O(s^2)) = \mathbb{E}_p[Z] + O(s)$, yielding $\lim_{s\to 0}\mathcal{R}_s(Z;p) = \mathbb{E}_p[Z]$. In $\mathcal{L}_{\mathrm{IRIS}}$, the first term has order $s_1 = \alpha\!-\!1 \to 0$ while the second has order $s_2 = \alpha \to 1$. Taking these limits:
\begin{equation}\label{eq:kl_limit}
\lim_{\alpha\to 1}\mathcal{L}_{\mathrm{IRIS}}(\theta;\alpha) = -\mathbb{E}_{p_{\mathrm{data}}}\!\left[r(\mathbf{x},\mathbf{y})\right] + \log\mathbb{E}_{p_{\theta_t}}\!\left[e^{r(\mathbf{x},\mathbf{y}')}\right].
\end{equation}
This is the negated Donsker-Varadhan representation~\cite{donsker1983asymptotic} of $\mathrm{KL}(p_{\mathrm{data}}\|p_{\theta_t})$ evaluated at the test function $f = r$. Since $r = \log(p_\theta/p_{\theta_t})$ and $p_\theta$ is a normalized distribution, the second term evaluates to $\log\int p_{\theta_t}\cdot(p_\theta/p_{\theta_t})\,d\mathbf{y} = \log 1 = 0$. The objective therefore reduces to $-\mathbb{E}_{p_{\mathrm{data}}}[\log p_\theta(\mathbf{y}|\mathbf{x})] + \mathrm{constant}$, recovering the negative log-likelihood used in supervised fine-tuning. This is the same KL regime as SPIN~\cite{chen2024self} and its variants~\cite{wang2026triplets,liao2026drift,alami2024regularization}, which approximate this objective through logistic losses on the reward gap $r(\mathbf{x},\mathbf{y})-r(\mathbf{x},\mathbf{y}')$.

Correspondingly, the importance weights in~\eqref{eq:weights} satisfy $\tilde{w}^+\!(\mathbf{y}) = e^{(\alpha-1)r}/\mathbb{E}[e^{(\alpha-1)r}] \to 1$ as $\alpha \to 1$, confirming that the gradient becomes a uniformly weighted likelihood update.

\subsubsection{$\chi^2$ Regime ($\alpha = 2$)}

Setting $\alpha = 2$ in~\eqref{eq:iris} and using $e^r = \rho = p_\theta/p_{\theta_t}$:
\begin{equation}\label{eq:chi2_obj}
\mathcal{L}_{\mathrm{IRIS}}(\theta;2) = -\log\mathbb{E}_{p_{\mathrm{data}}}\!\left[\rho\right] + \frac{1}{2}\log\mathbb{E}_{p_{\theta_t}}\!\left[\rho^2\right].
\end{equation}
We analyze each term. For the second term, expanding the expectation:
\begin{equation}\label{eq:chi2_expand}
\mathbb{E}_{p_{\theta_t}}\!\left[\rho^2\right] = \int p_{\theta_t}(\mathbf{y})\left(\frac{p_\theta(\mathbf{y})}{p_{\theta_t}(\mathbf{y})}\right)^{\!2}\,d\mathbf{y} = \int\frac{p_\theta(\mathbf{y})^2}{p_{\theta_t}(\mathbf{y})}\,d\mathbf{y} = 1 + \chi^2(p_\theta\|p_{\theta_t}),
\end{equation}
where the last equality follows from the definition $\chi^2(p\|q) = \int(p-q)^2/q = \int p^2/q - 1$~\cite{pearson1900criterion}. Therefore $\frac{1}{2}\log\mathbb{E}_{p_{\theta_t}}[\rho^2] = \frac{1}{2}\log(1+\chi^2(p_\theta\|p_{\theta_t}))$, which directly penalizes the $\chi^2$ divergence between the current policy and the opponent, confirming the connection to the bounded-reward formulations in SPIF~\cite{li2026your} and SPPO~\cite{wu2024sppo}. The importance weights at $\alpha = 2$ become $\tilde{w}^+\!\propto e^{r} = p_\theta/p_{\theta_t}$ and $\tilde{w}^-\!\propto e^{2r} = (p_\theta/p_{\theta_t})^2$, concentrating quadratically on synthetic responses that the model incorrectly favors.

\subsubsection{Hellinger Regime ($\alpha = 1/2$)}

Setting $\alpha = 1/2$ gives orders $\alpha\!-\!1 = -1/2$ and $\alpha = 1/2$ for the two tilted risks:
\begin{equation}\label{eq:hel_obj}
\mathcal{L}_{\mathrm{IRIS}}(\theta;\tfrac{1}{2}) = 2\log\mathbb{E}_{p_{\mathrm{data}}}\!\left[\rho^{-1/2}\right] + 2\log\mathbb{E}_{p_{\theta_t}}\!\left[\rho^{1/2}\right].
\end{equation}
The second expectation evaluates to the Bhattacharyya coefficient~\cite{bhattacharyya1943measure}:
\begin{equation}\label{eq:bhatt}
\mathbb{E}_{p_{\theta_t}}\!\left[\rho^{1/2}\right] = \int p_{\theta_t}(\mathbf{y})\left(\frac{p_\theta(\mathbf{y})}{p_{\theta_t}(\mathbf{y})}\right)^{\!1/2}\!\!d\mathbf{y} = \int\!\sqrt{p_\theta(\mathbf{y})\,p_{\theta_t}(\mathbf{y})}\,d\mathbf{y} = 1 - H^2(p_\theta,\,p_{\theta_t}),
\end{equation}
where $H^2(p,q) = 1 - \int\sqrt{pq}$ denotes the squared Hellinger distance~\cite{vanerven2014renyi}. The importance weights at $\alpha = 1/2$ are $\tilde{w}^+\!\propto e^{-r/2} = (p_{\theta_t}/p_\theta)^{1/2}$, which assign larger weight to annotated responses where the current model underperforms relative to the opponent, and $\tilde{w}^-\!\propto e^{r/2} = (p_\theta/p_{\theta_t})^{1/2}$, which mildly penalizes synthetic responses where the model overestimates. This inverse-exponential weighting induces a mode-covering effect that ensures the model does not ignore low-probability but high-quality real responses, a behavior absent from the KL and $\chi^2$ regimes where $\tilde{w}^+$ increases with $r$.

\begin{table}[t]
\centering
\caption{Wall-clock time (hours) for generation (\textit{Gen}) and training (\textit{Train}) at each iteration of IRIS, along with the adaptive order $\alpha_t$. All experiments use 8 NVIDIA H100 GPUs.}
\label{tab:cost}
\vspace{2pt}
\small
\begin{tabular}{cl ccccc}
\toprule
\multirow{2}{*}{Model} & \multirow{2}{*}{Phase} & \multicolumn{5}{c}{Iteration} \\
\cmidrule(lr){3-7}
& & Iter\,0 & Iter\,1 & Iter\,2 & Iter\,3 & Iter\,4 \\
\midrule
\multirow{3}{*}{Zephyr-7B} & $\alpha_t$ & 3.0 & 1.7 & 1.4 & 1.1 & 0.7 \\
& Gen & 0.81 & 0.81 & 0.81 & 0.81 & 0.81 \\
& Train & 3.31 & 3.27 & 3.22 & 3.19 & 3.17 \\
\midrule
\multirow{3}{*}{Qwen2.5-3B} & $\alpha_t$ & 2.5 & 1.8 & 1.4 & 1.2 & 0.9 \\
& Gen & 0.35 & 0.35 & 0.35 & 0.35 & 0.35 \\
& Train & 1.45 & 1.41 & 1.38 & 1.36 & 1.35 \\
\bottomrule
\end{tabular}
\end{table}

\section{More Experimental Details}\label{app:exp_details}

In this section, we provide implementation details, computational cost analysis, and full experimental results on Qwen2.5-3B-Instruct\cite{yang2025qwen3} that complement the main experiments in Section~\ref{sec:exp}.

\subsection{Implementation Details}\label{app:impl}

We build our codebase on the Alignment Handbook\cite{tunstall2024alignment} and use the Accelerate library\cite{gugger2022accelerate} for distributed training across 8 NVIDIA H100 GPUs with 80GB memory each. Following Chen et al.\cite{chen2024self} and Wang et al.\cite{wang2026triplets}, we employ RMSProp\cite{hinton2012neural} with no weight decay as the optimizer and set the maximum sequence length to 2048 tokens. The warmup ratio is 10\% of the total training steps, followed by a cosine learning rate schedule with a peak learning rate of $5 \times 10^{-7}$. We use a per-GPU batch size of 8, yielding a global batch size of 64. All training is conducted in bfloat16 mixed precision with DeepSpeed ZeRO Stage~3\cite{rajbhandari2020zero} and FlashAttention-2\cite{dao2024flashattention2} for memory efficiency. For synthetic response generation, we use the Accelerate library in distributed mode with a global batch size of 256. At each iteration, we generate one response per prompt using nucleus sampling with temperature $0.7$ and $\mathrm{top\text{-}}p = 0.9$, with a maximum generation length of 2048 tokens. Following SPIN\cite{chen2024self} and SPACE\cite{wang2025space}, we sample 50$k$ prompt-response pairs exclusively from the first round of interactions in UltraChat200k\cite{ding2023enhancing}. The prompts are formatted using the ChatML template consistent with each base model's instruction-tuning format.

\begin{table}[t]
\centering
\caption{Performance (\%) comparisons on Qwen2.5-3B-Instruct, where we report improvements ({\color[rgb]{0,0.5,0}green}) or degradations ({\color{red}red}) over the previous iteration. \textbf{Bold}: best result across all iterations.}
\label{tab:qwen}
\vspace{2pt}
\resizebox{\textwidth}{!}{%
\begin{tabular}{rc ccc ccc ccc c c}
\toprule
\multicolumn{2}{c}{\multirow{2}{*}{\textbf{Model}}} & \multicolumn{3}{c}{Math \& Logic} & \multicolumn{3}{c}{Multi-Domain Knowledge} & \multicolumn{3}{c}{Commonsense Reasoning} & IF & \multirow{2}{*}{\textbf{Avg}} \\
\cmidrule(lr){3-5} \cmidrule(lr){6-8} \cmidrule(lr){9-11} \cmidrule(lr){12-12}
& & GSM8K & MATH & MUSR & MMLU & MMLUP & GPQA & HellaSwag & WG & BBH & IFEval & \\
\midrule
\multicolumn{2}{c}{Qwen2.5-3B} & 67.84 & 34.63 & 75.72 & 62.16 & 24.57 & 26.43 & 71.56 & 61.38 & 50.85 & 26.73 & 50.19 \\
\multicolumn{2}{c}{SFT} & 71.43 & 36.28 & 77.15 & 64.30 & 24.62 & 27.18 & 72.14 & 63.27 & 51.73 & 33.82 & 52.19 \\
\midrule
\multirow{5}{*}{\rotatebox[origin=c]{90}{SPIN}}
& Iter0 & 73.27 & 38.62 & 78.14 & 67.48 & 24.32 & 28.00 & 73.45 & 64.83 & 51.08 & 26.58 & 52.58 \\
& Iter1 & 74.74 & 39.23 & 77.18 & 67.62 & 24.08 & 28.24 & 73.12 & 64.97 & 51.25 & 22.43 & 52.39\,{\scriptsize\color{red}($-$0.19)} \\
& Iter2 & 74.83 & 39.97 & 78.49 & 67.85 & 24.73 & 27.83 & 73.09 & 65.12 & 52.68 & 33.25 & \textbf{53.88}\,{\scriptsize\color[rgb]{0,0.5,0}($+$1.49)} \\
& Iter3 & 72.48 & 38.71 & 77.24 & 66.19 & 23.68 & 27.23 & 72.93 & 65.38 & 52.24 & 35.17 & 53.22\,{\scriptsize\color{red}($-$0.66)} \\
& Iter4 & 72.60 & 38.53 & 76.87 & 66.64 & 23.11 & 26.87 & 72.78 & 65.24 & 51.93 & 34.51 & 53.11\,{\scriptsize\color{red}($-$0.11)} \\
\midrule
\multirow{5}{*}{\rotatebox[origin=c]{90}{SPACE}}
& Iter0 & 72.78 & 38.84 & 77.26 & 66.50 & 25.24 & 28.15 & 73.12 & 64.93 & 52.23 & 28.47 & 52.75 \\
& Iter1 & 73.12 & 40.15 & 77.65 & 67.87 & 25.37 & 28.43 & 73.27 & 64.81 & 52.42 & 29.73 & 53.28\,{\scriptsize\color[rgb]{0,0.5,0}($+$0.53)} \\
& Iter2 & 73.38 & 40.49 & 78.07 & 68.03 & 25.48 & 28.57 & 73.19 & 65.05 & 52.68 & 31.40 & 53.63\,{\scriptsize\color[rgb]{0,0.5,0}($+$0.35)} \\
& Iter3 & 73.57 & 41.58 & 78.38 & 68.91 & 25.56 & 28.68 & 73.24 & 65.14 & 52.93 & 34.14 & 54.21\,{\scriptsize\color[rgb]{0,0.5,0}($+$0.58)} \\
& Iter4 & 73.84 & 41.68 & 78.50 & 68.85 & 25.63 & 28.81 & 73.31 & 64.93 & 53.18 & 36.52 & \textbf{54.52}\,{\scriptsize\color[rgb]{0,0.5,0}($+$0.31)} \\
\midrule
\multirow{5}{*}{\rotatebox[origin=c]{90}{T-SPIN}}
& Iter0 & 71.69 & 39.28 & 76.47 & 67.73 & 25.18 & 27.84 & 72.53 & 64.79 & 52.35 & 24.53 & 52.24 \\
& Iter1 & 72.37 & 39.43 & 77.12 & 68.28 & 25.42 & 28.56 & 72.84 & 65.03 & 52.72 & 38.49 & 54.03\,{\scriptsize\color[rgb]{0,0.5,0}($+$1.79)} \\
& Iter2 & 73.58 & 41.68 & 77.53 & 68.42 & 25.57 & 28.93 & 73.01 & 64.96 & 53.08 & 39.83 & 54.66\,{\scriptsize\color[rgb]{0,0.5,0}($+$0.63)} \\
& Iter3 & 75.42 & 41.90 & 78.14 & 68.98 & 25.73 & 29.38 & 73.18 & 65.04 & 53.73 & 42.64 & 55.41\,{\scriptsize\color[rgb]{0,0.5,0}($+$0.75)} \\
& Iter4 & 75.18 & 42.07 & 78.52 & 69.93 & 25.68 & 29.74 & 73.08 & 65.10 & 53.17 & 44.87 & \textbf{55.73}\,{\scriptsize\color[rgb]{0,0.5,0}($+$0.32)} \\
\midrule
\multirow{5}{*}{\rotatebox[origin=c]{90}{SPIF}}
& Iter0 & 70.41 & 38.03 & 78.63 & 66.58 & 24.81 & 28.49 & 73.70 & 64.65 & 51.56 & 25.38 & 52.22 \\
& Iter1 & 72.73 & 38.27 & 78.78 & 67.14 & 25.18 & 28.65 & 73.38 & 64.82 & 52.63 & 31.84 & 53.34\,{\scriptsize\color[rgb]{0,0.5,0}($+$1.12)} \\
& Iter2 & 71.56 & 40.43 & 79.08 & 67.61 & 25.47 & 28.87 & 73.18 & 64.98 & 52.68 & 38.28 & \textbf{54.21}\,{\scriptsize\color[rgb]{0,0.5,0}($+$0.87)} \\
& Iter3 & 72.83 & 39.35 & 78.52 & 65.24 & 25.22 & 28.62 & 73.05 & 64.84 & 52.37 & 37.85 & 53.79\,{\scriptsize\color{red}($-$0.42)} \\
& Iter4 & 72.14 & 39.22 & 78.19 & 66.87 & 24.98 & 28.38 & 72.92 & 64.73 & 52.12 & 37.43 & 53.70\,{\scriptsize\color{red}($-$0.09)} \\
\midrule
\rowcolor{red!6}
& Iter0 & 72.84 & 39.72 & 78.87 & 66.63 & 25.28 & 28.53 & 73.68 & 64.82 & 51.73 & 28.37 & 53.05 \\
\rowcolor{red!6}
& Iter1 & 74.85 & 40.27 & 79.13 & 66.90 & 25.87 & 29.43 & 73.42 & 65.14 & 52.28 & 37.62 & 54.49\,{\scriptsize\color[rgb]{0,0.5,0}($+$1.44)} \\
\rowcolor{red!6}
& Iter2 & 75.23 & 39.58 & 79.37 & 67.38 & 26.23 & 29.78 & 73.35 & 65.37 & 52.65 & 40.83 & 54.98\,{\scriptsize\color[rgb]{0,0.5,0}($+$0.49)} \\
\rowcolor{red!6}
& Iter3 & 75.48 & 40.83 & 79.65 & 68.72 & 26.49 & 29.91 & 73.42 & 65.53 & 53.37 & 43.51 & 55.69\,{\scriptsize\color[rgb]{0,0.5,0}($+$0.71)} \\
\rowcolor{red!6}
\multirow{-5}{*}{\rotatebox[origin=c]{90}{\textbf{IRIS}}}
& Iter4 & 75.72 & 41.05 & 79.83 & 68.93 & 26.60 & 30.08 & 73.48 & 65.68 & 53.93 & 46.47 & \textbf{56.18}\,{\scriptsize\color[rgb]{0,0.5,0}($+$0.49)} \\
\bottomrule
\end{tabular}%
}
\end{table}

\begin{figure}[h]
    \centering
    \includegraphics[width=\textwidth]{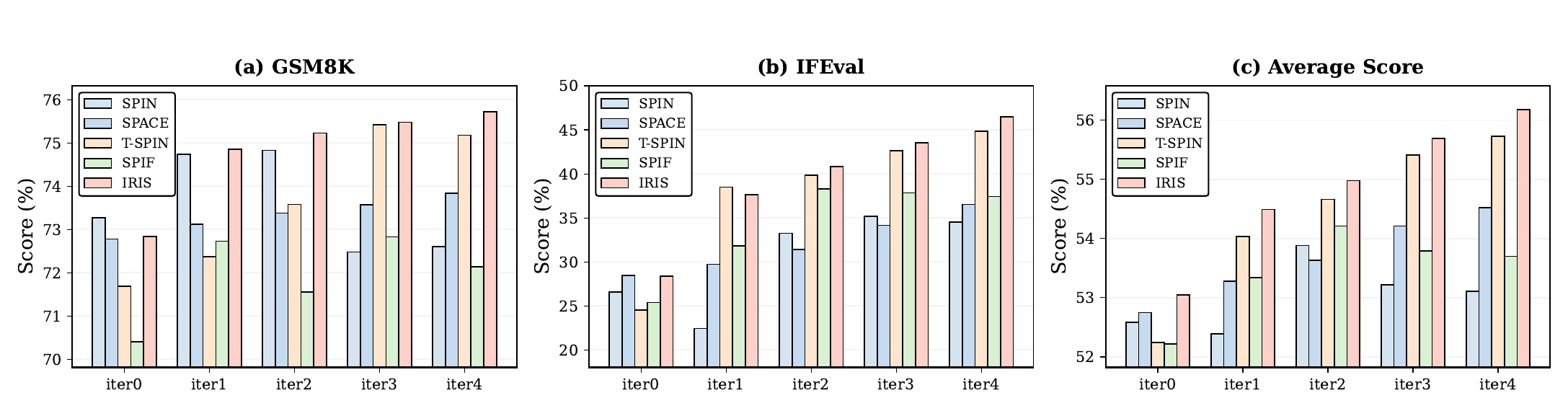}
    \caption{Performance comparison across self-play iterations on Qwen2.5-3B-Instruct for three representative metrics: (a) GSM8K, (b) IFEval, and (c) average score over all ten benchmarks. IRIS always achieves the highest scores and maintains consistent improvements throughout training.}
    \label{fig:qwen_comparison}
\end{figure}

\subsection{Computational Costs}\label{app:cost}

Table~\ref{tab:cost} reports the wall-clock time for generation and training at each iteration on both Zephyr-7B and Qwen2.5-3B-Instruct, together with the adaptive order $\alpha_t$ selected by the distributional gap feedback schedule in Eq.~\eqref{eq:adapt}. Generation costs remain constant across iterations at 0.81 hours for Zephyr-7B and 0.35 hours for Qwen2.5-3B, as the same number of synthetic responses (50$k$) are produced each time using the Accelerate library\cite{gugger2022accelerate} with distributed inference across 8 H100 GPUs.

Training time exhibits a mild but consistent decrease over iterations, dropping from 3.31 to 3.17 hours on Zephyr-7B and from 1.45 to 1.35 hours on Qwen2.5-3B-Instruct. This trend directly reflects the decay of $\alpha_t$ across iterations. In early iterations, larger $\alpha$ produces sharper exponential importance weights $\tilde{w}^{\pm} \propto \exp((\alpha\!-\!1)\,r)$ in Eq.~\eqref{eq:weights}, which increases the computational overhead of log-sum-exp stabilization and per-batch weight normalization. As training progresses and $\alpha_t$ decreases, the tilted risk terms in Eq.~\eqref{eq:iris} reduce toward simpler expectations with near-uniform weights, lowering this overhead. The total cost for five iterations is 20.2 hours for Zephyr-7B, and 8.7 hours for Qwen2.5-3B.

\subsection{Experimental Results on Qwen2.5-3B-Instruct}\label{app:qwen}

Table~\ref{tab:qwen} presents the performance of all methods on Qwen2.5-3B-Instruct across ten benchmarks over five self-play iterations. The overall trends are consistent with the Zephyr-7B results in Table~\ref{tab:main}. IRIS achieves the highest average score of 56.18\% at Iter~4, surpassing T-SPIN (55.73\%), SPACE (54.52\%), SPIF (54.21\%), and SPIN (53.88\%) by margins of 0.45, 1.66, 1.97, and 2.30 points, respectively. IRIS maintains monotonic improvement across all five iterations, whereas SPIN peaks at Iter~2 before declining, and SPIF similarly degrades after Iter~2. The gains are most pronounced on IFEval, where IRIS obtains a 19.74-point improvement over the base model, and on MMLU, where IRIS reaches 68.93\% compared to 66.64\% for SPIN. With only 50$k$ annotated samples trained for four iterations, IRIS surpasses SFT trained on the full 200$k$ dataset by 3.99 points, confirming that the adaptive divergence mechanism provides consistent benefits across model scales. Figure~\ref{fig:qwen_comparison} provides a visual comparison on GSM8K, IFEval, and average score across iterations.

\section{Generated Samples}\label{app:samples}

We provide a representative example to qualitatively illustrate the progressive evolution of responses generated by the model trained with IRIS across self-play iterations. Table~\ref{tab:samples} presents the annotated response and synthetic responses from the Zephyr-7B base model, SFT, and IRIS at each iteration (Iter~0 through Iter~4) for a given prompt.

\clearpage

\begin{table}[htbp]
\centering
\caption{Example responses from Zephyr-7B at different training stages.}
\label{tab:samples}
\begin{tabular*}{\textwidth}{@{}|>{\raggedright\arraybackslash}p{0.14\textwidth}|>{\raggedright\arraybackslash}p{0.83\textwidth}|@{}}
\hline
Prompt &
\parbox[t]{\linewidth}{\footnotesize\fontfamily{pcr}\selectfont Can you suggest a better approach for deleting configuration objects that would be faster than querying them individually?\newline\newline
The uninstall uses a technique of deleting configuration and syncing triggers to remove the triggers first so it can drop sym tables. It deletes configuration by querying all of the configuration objects (trigger routers, routers, group links, etc) and removing each one. The problem is that the query uses a select with all the columns known in the current version, which doesn't work on an older database. Change the approach to just delete all rows from configuration tables instead of querying them, which will be faster anyway.\vspace{3pt}} \\
\hline
Annotated Data &
\parbox[t]{\linewidth}{\footnotesize\fontfamily{pcr}\selectfont You could also consider using batch processing to delete configuration objects in larger groups, rather than individually. This would reduce the number of queries made and increase the speed of the deletion process. Additionally, you could use multithreading or parallel processing to further increase the speed of the deletion process. Finally, you could consider creating an index on the configuration table to speed up the query process, allowing for faster deletion.\vspace{3pt}} \\
\hline
Base Model &
\parbox[t]{\linewidth}{\footnotesize\fontfamily{pcr}\selectfont You can try to use a stored procedure or a cascading delete to remove configuration objects more efficiently. Another option is to restructure the database tables so that removing one parent record automatically removes all associated child records through ON DELETE CASCADE constraints.\vspace{3pt}} \\
\hline
SFT &
\parbox[t]{\linewidth}{\footnotesize\fontfamily{pcr}\selectfont Instead of querying each configuration object with a version-specific SELECT, you could execute direct DELETE statements on each configuration table. This removes the dependency on column names and is compatible with older database versions. Using TRUNCATE TABLE would also be faster for clearing entire tables during uninstall.\vspace{3pt}} \\
\hline
Iter\,0 &
\parbox[t]{\linewidth}{\footnotesize\fontfamily{pcr}\selectfont You could try grouping the deletion of configuration objects rather than handling them one at a time. Running a single DELETE statement per table would cut down on the number of queries to the database. This also gets around the column compatibility issue since you are not selecting specific columns from older schemas.\vspace{3pt}} \\
\hline
Iter\,1 &
\parbox[t]{\linewidth}{\footnotesize\fontfamily{pcr}\selectfont You could use batch processing to delete configuration objects in larger groups, rather than removing them individually. This would reduce the number of queries made and speed up the deletion process. Running the deletions in this way also avoids the problem of version-specific column names, since DELETE statements do not reference individual columns.\vspace{3pt}} \\
\hline
Iter\,2 &
\parbox[t]{\linewidth}{\footnotesize\fontfamily{pcr}\selectfont You could use batch processing to delete configuration objects in larger groups, rather than individually. This would reduce the number of queries made and increase the speed of the deletion process. You could also use parallel processing to delete from multiple configuration tables at the same time, which would further improve the overall deletion speed.\vspace{3pt}} \\
\hline
Iter\,3 &
\parbox[t]{\linewidth}{\footnotesize\fontfamily{pcr}\selectfont You could use batch processing to delete configuration objects in groups instead of one by one, which would reduce the overall number of queries and make the deletion faster. Multithreading could also be applied to process multiple configuration tables in parallel for additional speedup. It may also help to add an index on the configuration table to make the lookups faster when identifying which rows to remove.\vspace{3pt}} \\
\hline
Iter\,4 &
\parbox[t]{\linewidth}{\footnotesize\fontfamily{pcr}\selectfont You could consider using batch processing to delete configuration objects in larger groups, rather than individually. This would reduce the number of queries made and increase the speed of the deletion process. You could also use multithreading or parallel processing to further increase the deletion speed. Additionally, creating an index on the configuration table could speed up the query process and allow for faster deletion.\vspace{3pt}} \\
\hline
\end{tabular*}
\end{table}

\clearpage

\section{Pseudocode}\label{app:pseudocode}

In the following, we provide the PyTorch pseudocode for computing the IRIS loss~\eqref{eq:iris} and the adaptive order schedule~\eqref{eq:adapt}.

\begin{figure}[h]
\begin{lstlisting}[language=Python, basicstyle=\footnotesize\ttfamily, keywordstyle=\color{blue}, commentstyle=\color{green!50!black}, stringstyle=\color{red!70!black}, showstringspaces=false, frame=single, xleftmargin=1em, xrightmargin=1em, aboveskip=0.5em, belowskip=0.5em]
import torch
import torch.nn.functional as F

def iris_loss(policy_real_logps, policy_generated_logps,
              opponent_real_logps, opponent_generated_logps, alpha):
    """
    Compute the IRIS loss (Eq. 5).
    Args:
        policy_real_logps: Policy log probs on real data
        policy_generated_logps: Policy log probs on synthetic data
        opponent_real_logps: Opponent log probs on real data
        opponent_generated_logps: Opponent log probs on synthetic data
        alpha: Renyi order (alpha > 0, alpha != 1)
    Returns:
        torch.Tensor: scalar IRIS loss
    """
    # Log-ratio rewards: r = log p_theta - log p_{theta_t}
    r_real = policy_real_logps - opponent_real_logps
    r_gen = policy_generated_logps - opponent_generated_logps

    # Term 1: tilted reward on real data (log-sum-exp stabilized)
    scaled_real = (alpha - 1.0) * r_real
    max_real = scaled_real.detach().max()
    term_real = -1.0 / (alpha - 1.0) * (
        torch.log(torch.mean(torch.exp(scaled_real - max_real))) + max_real)

    # Term 2: tilted penalty on synthetic data
    scaled_gen = alpha * r_gen
    max_gen = scaled_gen.detach().max()
    term_gen = 1.0 / alpha * (
        torch.log(torch.mean(torch.exp(scaled_gen - max_gen))) + max_gen)

    return term_real + term_gen

def compute_alpha(real_rewards, generated_rewards,
                  c=0.5, alpha_min=0.5, alpha_max=3.0):
    """
    Adaptive order via distributional gap feedback (Eq. 7).
    """
    gap = real_rewards.mean() - generated_rewards.mean()
    alpha = 1.0 + c * gap
    return max(alpha_min, min(alpha_max, alpha))
\end{lstlisting}
\end{figure}





\end{document}